\documentclass[conference,10pt,twocolumn]{IEEEtran}

\usepackage{geometry}
\usepackage{graphicx}
\usepackage{amsmath,amssymb,amsthm}
\usepackage{booktabs}
\usepackage{multirow}
\usepackage{tikz}
\usetikzlibrary{shapes,arrows,positioning,calc,fit,backgrounds,%
decorations.pathreplacing,patterns,shapes.geometric,arrows.meta}
\usepackage{pgfplots}
\pgfplotsset{compat=1.18}
\usepgfplotslibrary{fillbetween}
\usepackage{xcolor}
\usepackage{enumitem}
\usepackage{float}
\usepackage{array}
\usepackage{tabularx}
\usepackage{cite}
\usepackage{stfloats}
\usepackage{balance}
\PassOptionsToPackage{hyphens}{url}
\usepackage[hidelinks,breaklinks]{hyperref}

\definecolor{sinkColor}{RGB}{150, 75, 70}      % first token sink
\definecolor{gateColor}{RGB}{70, 100, 140}     % gating
\definecolor{kdaColor}{RGB}{80, 110, 90}       % linear attention
\definecolor{depthColor}{RGB}{110, 85, 130}    % depth mixing
\definecolor{inkColor}{RGB}{95, 95, 95}        % neutral rules and text
\definecolor{warnColor}{RGB}{160, 130, 70}     % thresholds
\definecolor{paleFill}{RGB}{247, 247, 247}

\tikzset{
  layerbox/.style={rectangle, rounded corners=2pt, draw=inkColor!75,
    fill=white, minimum width=2.0cm, minimum height=0.52cm,
    font=\scriptsize, line width=0.55pt, inner sep=2pt},
  kdabox/.style={layerbox, draw=kdaColor!85, fill=kdaColor!8},
  mlabox/.style={layerbox, draw=gateColor!85, fill=gateColor!8},
  stagebox/.style={rectangle, rounded corners=4pt, draw=#1!55,
    fill=#1!4, line width=0.7pt},
  probe/.style={rectangle, rounded corners=2pt, draw=inkColor!70,
    fill=paleFill, minimum width=2.5cm, minimum height=0.55cm,
    font=\scriptsize, line width=0.55pt, inner sep=3pt},
  flow/.style={->, >=Stealth, line width=0.7pt, inkColor!85},
  thinflow/.style={->, >=Stealth, line width=0.5pt, inkColor!65},
  tag/.style={font=\scriptsize, text=inkColor},
  tinytag/.style={font=\tiny, text=inkColor},
}

\newcommand{\sinkmass}{\ensuremath{\sigma}}
\newcommand{\recgap}{\ensuremath{\Delta_{r}}}
\newcommand{\maxact}{\ensuremath{\mu}}
\newcommand{\R}{\mathbb{R}}
\newcommand{\E}{\mathbb{E}}

\newtheorem{definition}{Definition}

\title{Do New Attention Mechanisms Actually Fix Attention Sinks\\ at
Million-Token Context?}

\author{
\IEEEauthorblockN{Sara Rizwan\quad Samaanah Abdus Salam\quad Mohammed Mudassir Uddin}
\IEEEauthorblockA{
\textit{Shadan Women's College of Engineering and Technology}\\
\textit{Hyderabad, Telangana, India}\\
\textit{Muffakham Jah College Of Engineering \& Technology}\\
\textit{Hyderabad, Telangana, India}\\
\{sara.rizwan014@gmail.com,\; samaanah345@gmail.com,\; mohd.mudassiruddin7@gmail.com\}
}
}

\begin{document}

\maketitle

% ============================================
% ABSTRACT
% ============================================
\section*{Abstract}

Long context language models now advertise windows of one million tokens,
but two habits limit how much of that window is used. Attention heads
with nothing useful to read still spend their budget on the first token,
which is called the attention sink, and where a fact sits in the context
changes whether the model finds it. Gated attention cut first token
attention from 46.7 percent to 4.8 percent at NeurIPS 2025, and Kimi K3
pairs that idea with Kimi Delta Attention and Attention Residuals behind
a one million token window, eight times past the range where these
diagnostics have been reported. This paper asks whether the fix survives
that jump. We build SinkProbe, a suite that measures sink mass, massive
activation, position resolved recall and the recency gap, and apply it to
four small models that differ only in how they mix tokens and depth.
Three results follow. The training objective produces the sink, not the
architecture. Gating did not reproduce its published effect at our scale.
Sink mass, activations and position bias moved independently. Code, data
and the measurement protocol are released at
\url{https://github.com/sararizwan7/Attention-Mechanisms-in-1M-Context-Window}

\vspace{0.5em}
\noindent\textbf{Keywords}. Attention sink, long context language models,
linear attention, recency bias, model evaluation.

% ============================================
% 1. INTRODUCTION
% ============================================
\section{Introduction}

Ask a language model to remember the first sentence you gave it five
hundred thousand words ago and you learn something about how it spends
its attention. The model has read every one of those words. It has kept
them inside its context window. Yet the answer often comes back vague,
wrong, or built out of the last few pages instead of the first one. The
window was long. The memory was not.

This gap between window length and usable memory is the subject of this
paper. It has two named causes, and they are usually studied apart.

\subsection{The First Habit, Attention Sinks}

Softmax attention \cite{vaswani2017attention} forces every head to spend
a budget of exactly one across the tokens it can see. When a head finds
nothing worth reading, it cannot spend nothing. It has to put the budget
somewhere. In almost every trained transformer that somewhere is the
first position, which normally holds a beginning of sequence marker with
no meaning of its own \cite{xiao2024streaming}. Xiao et al. named the pattern the attention
sink and showed it is so load bearing that evicting those first few
tokens from the cache makes a streaming model collapse.

The number is larger than the name suggests. Qiu et al. measured a fifteen
billion parameter model sending 46.7 percent of its attention to the
first token, averaged across all layers, with one layer sending 83
percent \cite{qiu2025gated}. Almost half of the model's reading capacity
was being spent on a marker. A budget spent there is a budget not spent
on the sentence the user cared about.

\subsection{The Second Habit, Reading the Window Unevenly}

The second habit is easier to notice and harder to measure. Where in the
context a fact sits changes how likely the model is to find it. Liu et
al. called this being lost in the middle and showed a clear U shape in
accuracy against evidence position, with the two ends of the context
answered well and the centre answered poorly \cite{liu2024lost}. Baker et
al. showed the shape survives when several separate facts must be
combined \cite{baker2024middle}, and Hengle et al. showed it grows worse
outside English \cite{hengle2024mlneedle}. In large models on natural
text the stronger of the two ends is usually the recent one, which gives
the user facing symptom from the opening paragraph. The model quotes the
last page back perfectly and forgets the first.

We measure this habit with a single signed number, recall near the
question minus recall at the opening, and we call it the recency gap. The
name follows the usual direction. The sign is not assumed, and
Section~VI reports a case where it runs the other way.

\subsection{Why the Existing Evidence Stops Short}

Two recent lines of work attack these habits inside the architecture
rather than inside the prompt.

Gated attention places a small input dependent gate on the output of
scaled dot product attention \cite{qiu2025gated}. The gate gives a head
a way to output nothing, so a head with nothing to say no longer needs a
sink to absorb its budget. First token attention fell from 46.7 percent
to 4.8 percent and the largest hidden state activation fell from about
1053 to about 94 in the same comparison. The paper won a best paper award
and the mechanism has shipped in production models.

Kimi K3 takes a different route to the same goal and goes much further on
length \cite{kimi2026k3}. Most of its layers are replaced by Kimi Delta
Attention, a linear layer that carries a fixed size recurrent state with a
channel wise forget gate, so cost stops growing with sequence length. Its
remaining global layers are gated multi head latent attention
\cite{deepseek2024v2}, which already contains the gate from the NeurIPS
work. Its residual connections are replaced by Attention Residuals, which
let a layer attend over the outputs of earlier layers instead of pulling
them out of one crowded running sum \cite{kimi2026attnres}. Its feed
forward path activates 16 of 896 experts per token \cite{shazeer2017moe},
which is what keeps a 2.8 trillion parameter model affordable to run. It
drops positional encoding entirely, so nothing has to be rescaled when
the window is extended \cite{peng2023yarn}, and it advertises a context
window of one million tokens. Its predecessor at one trillion parameters
used softmax attention in every layer \cite{kimi2025k2}, so the change is
recent and deliberate.

The problem is that the two bodies of evidence do not meet. The gated
attention paper reports the sink diagnostics but stops at 128 thousand
tokens. The Kimi K3 report reaches one million tokens but reports
benchmark scores rather than the diagnostics. Nobody has run the first
measurement at the second length. So the strongest claim in long context
modelling today rests on scores rather than on evidence about the
mechanism those scores are supposed to come from.

\subsection{Research Question}

This leads to the question in the title, which we state precisely.

\begin{quote}
\textit{When a model replaces softmax attention with gated and linear
mechanisms and extends its window to one million tokens, does the
attention sink disappear, does the context start being read evenly, and
are those two outcomes the same event or two different ones?}
\end{quote}

The last part matters most. The sink and the uneven reading are usually
discussed as one problem. If they are two, then a fix for the first buys
nothing for the second, and every model in this family inherits the
second untouched.

\subsection{Contributions}

This paper makes five contributions, each stated with the limit of what
it supports.

\begin{itemize}[leftmargin=*]
  \item \textbf{SinkProbe, a four metric diagnostic suite.} Sink mass,
  massive activation, position resolved recall and the recency gap,
  defined in one place with an estimator and a confidence treatment for
  each. The definitions follow published practice \cite{qiu2025gated,
  sun2024massive, kamradt2023niah}. What is new is that they are
  collected, made comparable and made runnable.

  \item \textbf{A controlled ladder of four architectures.} Four models
  of about one million parameters that differ only in how they mix tokens
  and depth, with width, depth, heads, data, optimiser and seeds held
  fixed. They isolate mechanisms. They do not predict absolute scores at
  frontier scale, and Section VI reports a case where a mechanism known
  to work at scale does not appear at ours.

  \item \textbf{Evidence that the objective makes the sink.} Changing the
  training objective and nothing else removes almost all of the sink from
  an unchanged architecture. This explains where the pressure comes from.
  It does not measure how much survives at frontier scale.

  \item \textbf{A cache growth model for the Kimi K3 layer mix.} From the
  published layer counts, 69 of 93 layers carry a cache that does not
  grow with context. Two dimensions are unpublished and enter as declared
  assumptions. The layer fraction needs neither of them.

  \item \textbf{A protocol registered in advance.} Measurement points,
  sample sizes and numeric pass thresholds for the released weights,
  fixed before the runs. This is a commitment rather than a result, and
  its value is that it removes the freedom to decide afterwards what the
  numbers meant.
\end{itemize}

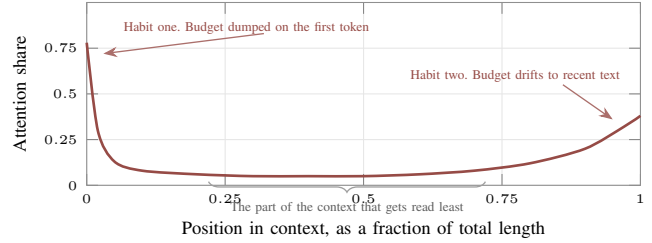
\begin{figure}[!t]
\centering
\begin{tikzpicture}
\begin{axis}[
    width=0.99\columnwidth,
    height=4.0cm,
    xlabel={Position in context, as a fraction of total length},
    ylabel={Attention share},
    xlabel style={font=\scriptsize},
    ylabel style={font=\scriptsize},
    xmin=0, xmax=1, ymin=0, ymax=1.0,
    xtick={0,0.25,0.5,0.75,1.0},
    ytick={0,0.25,0.5,0.75},
    x tick label style={font=\tiny},
    y tick label style={font=\tiny},
    grid=major, grid style={gray!20},
    axis line style={inkColor!70},
    clip=false,
]
% attention profile of an untreated stack
\addplot[sinkColor, line width=1.0pt, smooth] coordinates {
 (0.00,0.78) (0.02,0.30) (0.05,0.13) (0.10,0.08) (0.20,0.06)
 (0.30,0.05) (0.40,0.05) (0.50,0.05) (0.60,0.06) (0.70,0.08)
 (0.80,0.12) (0.90,0.20) (0.97,0.32) (1.00,0.38)
};
\node[tinytag, text=sinkColor, anchor=west] at (axis cs:0.05,0.86)
  {Habit one. Budget dumped on the first token};
\node[tinytag, text=sinkColor, anchor=east] at (axis cs:0.98,0.60)
  {Habit two. Budget drifts to recent text};
\draw[thinflow, sinkColor!80] (axis cs:0.32,0.83) -- (axis cs:0.03,0.72);
\draw[thinflow, sinkColor!80] (axis cs:0.86,0.55) -- (axis cs:0.95,0.36);
\draw[decorate, decoration={brace, amplitude=4pt, mirror}, inkColor!60,
      line width=0.5pt]
  (axis cs:0.22,0.02) -- (axis cs:0.72,0.02)
  node[midway, below=4pt, font=\tiny, text=inkColor]
  {The part of the context that gets read least};
\end{axis}
\end{tikzpicture}
\caption{The two habits that separate window length from usable memory,
drawn as one attention profile. Mass piles up on the first position
because a softmax head must spend a budget of one even when nothing is
worth reading, and mass drifts to the end because recent text is easiest
to use. The middle of the context is what the model reads least, which is
also where most of a long document lives. Curve shape follows the
patterns reported in \cite{xiao2024streaming} and \cite{liu2024lost}.}
\label{fig:problem}
\end{figure}

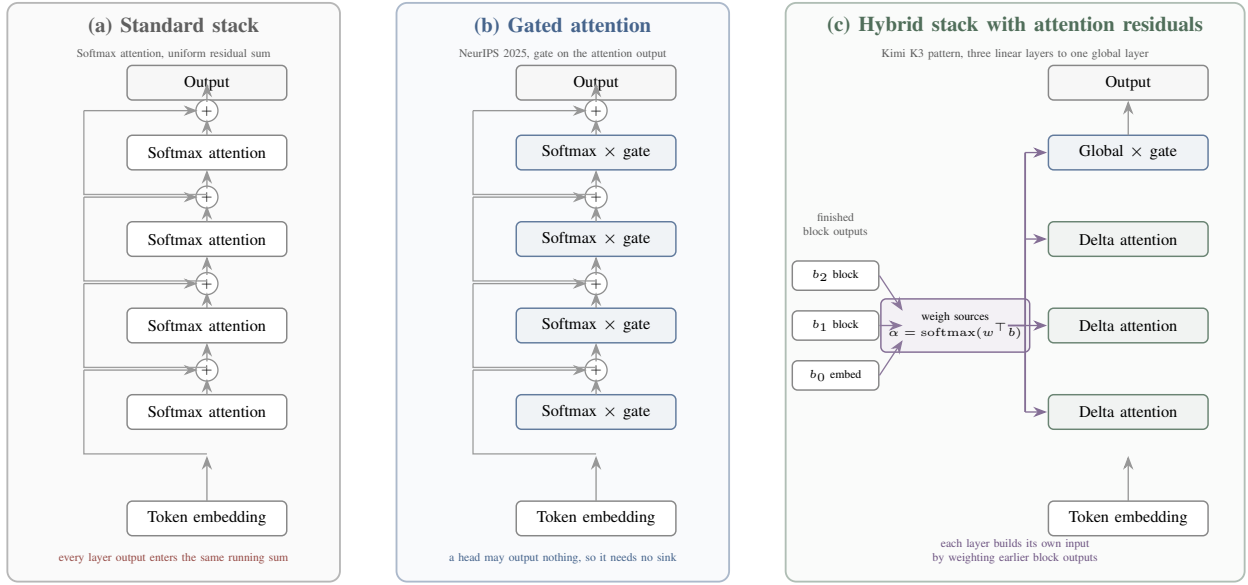
\begin{figure*}[!t]
\centering
\begin{tikzpicture}[scale=0.88, transform shape]

% ================= panel (a). standard stack =================
\begin{scope}[shift={(0,0)}]
  \draw[rounded corners=4pt, draw=inkColor!45, fill=inkColor!3, line width=0.7pt]
    (0.20,-2.55) rectangle (5.20,6.15);
  \node[font=\small\bfseries, text=inkColor] at (2.70,5.80) {(a) Standard stack};
  \node[tinytag] at (2.70,5.38) {Softmax attention, uniform residual sum};

  \node[layerbox, minimum width=2.4cm] at (3.20,-1.60) {Token embedding};
  \node[layerbox, minimum width=2.4cm] at (3.20,0.00) {Softmax attention};
  \node[layerbox, minimum width=2.4cm] at (3.20,1.30) {Softmax attention};
  \node[layerbox, minimum width=2.4cm] at (3.20,2.60) {Softmax attention};
  \node[layerbox, minimum width=2.4cm] at (3.20,3.90) {Softmax attention};
  \node[layerbox, minimum width=2.4cm, fill=paleFill] at (3.20,4.95) {Output};

  \draw[thinflow] (3.20,-1.34) -- (3.20,-0.66);
  \foreach \y in {0.00, 1.30, 2.60, 3.90} {
    \node[circle, draw=inkColor!70, fill=white, inner sep=1.0pt,
          font=\tiny, line width=0.5pt] at (3.20,\y+0.63) {$+$};
    \draw[thinflow] (3.20,\y+0.26) -- (3.20,\y+0.50);
    \draw[line width=0.5pt, inkColor!60]
      (3.20,\y-0.63) -- (1.35,\y-0.63) -- (1.35,\y+0.63);
    \draw[thinflow] (1.35,\y+0.63) -- (3.06,\y+0.63);
    \draw[thinflow] (3.20,\y+0.76) -- (3.20,\y+1.04);
  }
  \node[tinytag, text=sinkColor, align=center] at (2.70,-2.20)
    {every layer output enters the same running sum};
\end{scope}

% ================= panel (b). gated stack =================
\begin{scope}[shift={(5.85,0)}]
  \draw[rounded corners=4pt, draw=gateColor!45, fill=gateColor!3, line width=0.7pt]
    (0.20,-2.55) rectangle (5.20,6.15);
  \node[font=\small\bfseries, text=gateColor] at (2.70,5.80) {(b) Gated attention};
  \node[tinytag] at (2.70,5.38) {NeurIPS 2025, gate on the attention output};

  \node[layerbox, minimum width=2.4cm] at (3.20,-1.60) {Token embedding};
  \node[mlabox, minimum width=2.4cm] at (3.20,0.00) {Softmax $\times$ gate};
  \node[mlabox, minimum width=2.4cm] at (3.20,1.30) {Softmax $\times$ gate};
  \node[mlabox, minimum width=2.4cm] at (3.20,2.60) {Softmax $\times$ gate};
  \node[mlabox, minimum width=2.4cm] at (3.20,3.90) {Softmax $\times$ gate};
  \node[layerbox, minimum width=2.4cm, fill=paleFill] at (3.20,4.95) {Output};

  \draw[thinflow] (3.20,-1.34) -- (3.20,-0.66);
  \foreach \y in {0.00, 1.30, 2.60, 3.90} {
    \node[circle, draw=inkColor!70, fill=white, inner sep=1.0pt,
          font=\tiny, line width=0.5pt] at (3.20,\y+0.63) {$+$};
    \draw[thinflow] (3.20,\y+0.26) -- (3.20,\y+0.50);
    \draw[line width=0.5pt, inkColor!60]
      (3.20,\y-0.63) -- (1.35,\y-0.63) -- (1.35,\y+0.63);
    \draw[thinflow] (1.35,\y+0.63) -- (3.06,\y+0.63);
    \draw[thinflow] (3.20,\y+0.76) -- (3.20,\y+1.04);
  }
  \node[tinytag, text=gateColor, align=center] at (2.70,-2.20)
    {a head may output nothing, so it needs no sink};
\end{scope}

% ================= panel (c). hybrid stack with attention residuals =========
\begin{scope}[shift={(11.70,0)}]
  \draw[rounded corners=4pt, draw=kdaColor!45, fill=kdaColor!3, line width=0.7pt]
    (0.20,-2.55) rectangle (7.10,6.15);
  \node[font=\small\bfseries, text=kdaColor] at (3.65,5.80)
    {(c) Hybrid stack with attention residuals};
  \node[tinytag] at (3.65,5.38)
    {Kimi K3 pattern, three linear layers to one global layer};

  \node[layerbox, minimum width=2.4cm] at (5.35,-1.60) {Token embedding};
  \node[kdabox, minimum width=2.4cm] at (5.35,0.00) {Delta attention};
  \node[kdabox, minimum width=2.4cm] at (5.35,1.30) {Delta attention};
  \node[kdabox, minimum width=2.4cm] at (5.35,2.60) {Delta attention};
  \node[mlabox, minimum width=2.4cm] at (5.35,3.90) {Global $\times$ gate};
  \node[layerbox, minimum width=2.4cm, fill=paleFill] at (5.35,4.95) {Output};

  % sources that a layer may read from
  \node[layerbox, minimum width=1.30cm, minimum height=0.44cm, font=\tiny]
    at (0.95,0.55) {$b_0$ embed};
  \node[layerbox, minimum width=1.30cm, minimum height=0.44cm, font=\tiny]
    at (0.95,1.30) {$b_1$ block};
  \node[layerbox, minimum width=1.30cm, minimum height=0.44cm, font=\tiny]
    at (0.95,2.05) {$b_2$ block};
  \node[tinytag, align=center] at (0.95,2.80) {finished\\block outputs};

  \node[rectangle, rounded corners=2pt, draw=depthColor!85, fill=depthColor!8,
        minimum width=1.55cm, minimum height=0.80cm, font=\tiny, align=center,
        line width=0.55pt] at (2.75,1.30)
    {weigh sources\\$\alpha=\mathrm{softmax}(w^{\top}b)$};

  \draw[thinflow, depthColor!80] (1.61,0.55) -- (1.96,1.08);
  \draw[thinflow, depthColor!80] (1.61,1.30) -- (1.96,1.30);
  \draw[thinflow, depthColor!80] (1.61,2.05) -- (1.96,1.52);

  % bus that feeds every layer input
  \draw[line width=0.7pt, depthColor!85] (3.54,1.30) -- (3.80,1.30);
  \draw[line width=0.7pt, depthColor!85] (3.80,0.00) -- (3.80,3.90);
  \foreach \y in {0.00, 1.30, 2.60, 3.90} {
    \draw[thinflow, depthColor!85] (3.80,\y) -- (4.13,\y);
  }
  \draw[thinflow] (5.35,-1.34) -- (5.35,-0.66);
  \draw[thinflow] (5.35,4.16) -- (5.35,4.68);
  \node[tinytag, text=depthColor, align=center] at (3.65,-2.10)
    {each layer builds its own input\\by weighting earlier block outputs};
\end{scope}

\end{tikzpicture}
\caption{The three token and depth mixing rules compared in this paper.
In (a) every layer output is added into one running sum with unit weight,
and every layer reads through softmax attention, which is where the sink
forms. In (b) an input dependent gate on the attention output lets a head
emit nothing, which removes the reason to hold a sink \cite{qiu2025gated}.
In (c) three linear delta attention layers carry a fixed size recurrent
state and one gated global layer preserves full range interaction, while
attention residuals replace the running sum with a learned weighting over
earlier block outputs \cite{kimi2026k3, kimi2026attnres}. Kimi K3 repeats
the block shown in (c) throughout a 93 layer backbone, giving 69 linear
layers and 24 global layers.}
\label{fig:arch}
\end{figure*}

Figure~\ref{fig:problem} shows the two habits as one picture, and
Figure~\ref{fig:arch} shows the three architectures this paper compares.

% ============================================
% FIGURE. THREE ARCHITECTURES
% ============================================

% ============================================
% 2. RELATED WORK
% ============================================
\section{Related Work and the Gap it Leaves}

Work on this problem falls into five groups. We take each in turn and say
what it settled and what it left open, because the gap this paper fills
sits between two of them rather than inside any one.

\subsection{Finding and Explaining the Sink}

Xiao et al. named the attention sink while trying to make streaming
inference cheap \cite{xiao2024streaming}. Their finding was blunt. A
sliding window cache works until the first few tokens fall out of it, at
which point perplexity jumps by orders of magnitude, and keeping four
initial tokens pinned in the cache restores it. The tokens carry almost
no semantic content, so their value is structural rather than
informational.

Sun et al. connected the pattern to massive activations, a small number
of hidden state coordinates whose magnitude runs thousands of times above
the rest \cite{sun2024massive}. Gu et al. traced when the sink appears
during pre-training and showed it depends on optimisation and data rather
than on any single design choice \cite{gu2024sink}. Two further studies
argue about what the sink is for. One reads it as a null position that a
head uses when no token deserves attention, and the other reads it as a
staging area where the model parks information it will collect later
\cite{singular2025anchor, catch2025release}. Both readings predict the
same measurable signature, which is why a measurement rather than an
argument is what settles the question for a given model.

\subsection{Removing the Sink by Gating}

Qiu et al. showed that a small input dependent gate on the output of
scaled dot product attention removes the sink rather than working around
it \cite{qiu2025gated}. Their ablation is careful. A gate that is head
specific and query dependent cuts first token attention from 46.7 percent
to 4.8 percent and cuts the mean largest activation from about 1053 to
about 94. A gate that shares one score across heads cuts activations but
leaves the sink at 30.1 percent, which shows the two effects are
separable. Gating also helped when the context was extended past the
training length, where the gated model held 58.8 on the RULER benchmark
at 128 thousand tokens against 31.7 for the ungated baseline. A follow up
line combines the gate with sparse attention and reports the same
direction \cite{gatedsparse2026}, and a separate line reaches the same
end by replacing softmax with a function that is allowed to return zero
\cite{zuhri2025softpick}. Both support the same reading. The sink is a
consequence of forcing a head to spend a fixed budget, and any change
that lets a head spend less removes it.

Two limits matter here. The evidence covers dense softmax models up to
128 thousand tokens, and the model studied has fifteen billion total
parameters. Neither the architecture nor the length matches what is now
being deployed.

\subsection{Moving Information Across Depth}

A residual stream \cite{he2016residual} adds every layer output with the
same unit weight, and under the pre norm placement that modern models use
\cite{xiong2020prenorm} the hidden state magnitude grows with depth, so
the share belonging to any one layer shrinks. The Kimi team call this dilution and replace the sum
with softmax attention over previous layer outputs, which they name
Attention Residuals \cite{kimi2026attnres}. To keep memory bounded they
group layers into blocks and attend over block level representations
instead of individual layers, which drops the overhead from linear in
depth to linear in block count. A related proposal attends over the
difference between consecutive layers rather than over their outputs
\cite{deltaattnres2026}. This line targets depth. It says nothing about
where attention lands along the sequence, which is a separate axis.

\subsection{Linear Attention and Fixed Size State}

Linear attention replaces the growing key and value cache with a
recurrent state of fixed size, which removes the term that makes long
context expensive. Read as a fast weight memory \cite{schlag2021fast},
plain linear attention loses accuracy because the state has no way to
overwrite what it holds, so later work adds a delta rule that edits the
state in place \cite{yang2024deltanet} and a learned forget gate that
lets each channel decay at its own rate \cite{yang2024gla}. The competing
approach keeps softmax attention and makes it cheaper through better
kernels \cite{dao2022flash}, which lowers the constant but leaves the
growth in place. Kimi Delta Attention combines both, bounds the decay
from below so that the chunked kernel stays stable, and pairs three such
layers with one gated global layer \cite{kimi2025linear, kimi2026k3}.
Because the decay itself encodes order, the model needs no positional
encoding, and so nothing has to be retuned when the window is extended.

\subsection{Measuring Long Context Behaviour}

The needle in a haystack test plants a fact at a controlled depth and
asks for it back \cite{kamradt2023niah}. It is simple enough that strong
models saturate it, which is why LongBench and LongBench v2 add realistic
multi document tasks and why RULER varies the number and type of needles
\cite{bai2024longbench, bai2025longbenchv2, hsieh2024ruler}. Liu et al.
established the U shaped position curve \cite{liu2024lost}, Baker et al.
extended it to multi hop questions \cite{baker2024middle}, and Hengle et
al. showed the drop is larger for languages other than English
\cite{hengle2024mlneedle}. These benchmarks report what a model scores.
They do not report where the attention went, so a score alone cannot say
which of the two habits caused a failure.

\subsection{Position of This Work}

Table~\ref{tab:related} lays the five groups side by side. Reading down
the last two columns shows the gap. The diagnostics exist and stop at 128
thousand tokens. The million token architecture exists and reports no
diagnostics. This paper joins the two, and it treats the sink and the
recency gap as two measurements rather than one, because the ablation in
\cite{qiu2025gated} already showed that the underlying effects come
apart.

\begin{table}[!t]
\centering
\caption{Where each line of work stops. The last two columns are the ones
that leave a gap.}
\label{tab:related}
\footnotesize
\renewcommand{\arraystretch}{1.15}
\begin{tabular}{@{}>{\raggedright\arraybackslash}p{2.35cm}>{\raggedright\arraybackslash}p{1.55cm}cc@{}}
\toprule
\textbf{Line of work} & \textbf{Main idea} & \textbf{Reports} &
\textbf{Longest} \\
 & & \textbf{sink data} & \textbf{context} \\
\midrule
Streaming sinks \cite{xiao2024streaming} & Pin first tokens in cache &
Yes & 4M stream \\
Massive activations \cite{sun2024massive} & Outlier hidden units & Yes &
4K \\
Gated attention \cite{qiu2025gated} & Gate the attention output & Yes &
128K \\
Attention residuals \cite{kimi2026attnres} & Attend over depth & No &
Not stated \\
Linear attention \cite{yang2024gla, kimi2025linear} & Fixed size
recurrent state & No & 1M \\
Long context suites \cite{bai2025longbenchv2, hsieh2024ruler} & Score
retrieval and reasoning & No & 2M words \\
Kimi K3 \cite{kimi2026k3} & Hybrid, gated, no position encoding & No &
1M \\
\midrule
\textbf{This work} & \textbf{Sink diagnostics on the hybrid design} &
\textbf{Yes} & \textbf{1M} \\
\bottomrule
\end{tabular}
\end{table}

% ============================================
% 3. BACKGROUND AND PROBLEM FORMULATION
% ============================================
\section{Background and Problem Formulation}

This section fixes the notation used for the rest of the paper and turns
the two habits from Section I into quantities that can be measured. Every
symbol appears in Table~\ref{tab:notation}.

\subsection{Notation}

\begin{table}[!t]
\centering
\caption{Symbols used throughout the paper.}
\label{tab:notation}
\footnotesize
\renewcommand{\arraystretch}{1.15}
\begin{tabular}{@{}ll@{}}
\toprule
\textbf{Symbol} & \textbf{Meaning} \\
\midrule
$T$ & Context length in tokens \\
$L$ & Number of layers in the stack \\
$H$ & Number of attention heads per layer \\
$\mathcal{G}$ & Set of layers that use softmax attention \\
$A^{(\ell,h)}_{t,s}$ & Attention from query $t$ to key $s$, layer $\ell$,
head $h$ \\
$h_\ell \in \R^{d}$ & Hidden state entering layer $\ell$ \\
$\sinkmass$ & Sink mass, attention share on position zero \\
$\maxact$ & Massive activation, largest absolute hidden value \\
$\mathcal{H}$ & Attention entropy in nats \\
$p(d)$ & Recall when the answer sits at depth $d$ \\
$\recgap$ & Recency gap, late recall minus early recall \\
$S_t \in \R^{d_k \times d_v}$ & Recurrent state of a linear layer \\
$a_t$ & Channel wise decay applied to that state \\
$b_n$ & Representation summarising block $n$ \\
$\alpha_i^{\ell}$ & Weight layer $\ell$ places on source $i$ \\
\bottomrule
\end{tabular}
\end{table}

\subsection{Sink Mass}

A softmax head produces a distribution over the positions it can see, so
its weights sum to one whether or not any of those positions is useful.
Sink mass measures how much of that mandatory budget lands on the first
position.

\begin{definition}[Sink mass]
For a stack with softmax layers $\mathcal{G}$ evaluated on a context of
length $T$,
\begin{equation}
\sinkmass \;=\;
\frac{1}{|\mathcal{G}|\,H\,(T-1)}
\sum_{\ell \in \mathcal{G}} \sum_{h=1}^{H} \sum_{t=2}^{T}
A^{(\ell,h)}_{t,1} .
\label{eq:sink}
\end{equation}
\end{definition}

The first query position is skipped because it can only attend to itself,
which would inflate the estimate by a fixed amount. A stack that reads its
context evenly would score close to $1/T$, so at $T=160$ an untreated
value near $0.5$ means roughly eighty times more attention than an even
split would give. We also report the single worst layer, because the
average hides the concentration that Qiu et al. observed at one layer of
their model \cite{qiu2025gated}.

\subsection{Massive Activation}

Sinks travel with a handful of hidden state coordinates that grow far
beyond the rest \cite{sun2024massive}. We track the largest absolute
value entering each layer and average it over the stack,
\begin{equation}
\maxact \;=\; \frac{1}{L+1}\sum_{\ell=0}^{L}
\max_{t, j} \; \bigl| h_{\ell}[t, j] \bigr| .
\label{eq:act}
\end{equation}
This quantity is worth reporting separately because the ablation in
\cite{qiu2025gated} showed that a change can shrink $\maxact$ without
shrinking $\sinkmass$. Treating the two as one number would have hidden
that result.

\subsection{Attention Entropy}

Entropy says whether attention is spread or concentrated without
committing to a position,
\begin{equation}
\mathcal{H} \;=\;
-\frac{1}{|\mathcal{G}|\,H\,(T-1)}
\sum_{\ell, h, t} \sum_{s \le t}
A^{(\ell,h)}_{t,s} \log A^{(\ell,h)}_{t,s} .
\label{eq:entropy}
\end{equation}
A stack that dumps its budget on one position has entropy near zero even
when that position is not the first one, so entropy catches sinks that
move.

\subsection{Position Resolved Recall and the Recency Gap}

Let $p(d)$ be the probability that the model answers correctly when the
evidence sits at depth $d \in [0,1]$, where $d=0$ is the opening of the
context and $d=1$ is the token just before the question. The recency gap
compares the two ends,
\begin{equation}
\recgap \;=\;
\E_{d \in (0.75,\,1]}\bigl[p(d)\bigr] \;-\;
\E_{d \in [0,\,0.25]}\bigl[p(d)\bigr] .
\label{eq:recency}
\end{equation}
A positive $\recgap$ means the model prefers recent evidence. A value near
zero means the context is read evenly, which is what a long window is
supposed to deliver. The quantity is reported with a Wilson score
interval, since it is a difference of two binomial proportions and the
normal approximation is unreliable near the ends of the range
\cite{wilson1927}.

\subsection{What Makes Length Hard}

Three separate pressures grow with $T$, and they are worth separating
because different mechanisms address different ones.

\begin{itemize}[leftmargin=*]
  \item \textbf{Cost.} Softmax attention costs $O(T^2)$ time and its cache
  costs $O(T)$ memory per layer. At one million tokens the cache alone
  dominates the weights for most models, which is the pressure that
  linear attention removes.
  \item \textbf{Dilution along the sequence.} The budget of one is shared
  among more positions, so the share available to any single useful token
  falls as $T$ grows. A fixed sink cost of half the budget hurts more at
  one million tokens than at four thousand.
  \item \textbf{Dilution along the depth.} A uniform residual sum lets
  hidden state magnitude grow with depth, so an early layer contributes a
  smaller and smaller fraction of what a late layer reads
  \cite{kimi2026attnres}. This pressure is independent of $T$ but
  interacts with it, since long range evidence usually enters early.
\end{itemize}

\subsection{Problem Statement}

Given a model $M$ and a context length $T$, we want the four numbers
$\sinkmass(M,T)$, $\maxact(M,T)$, $p(d\,;M,T)$ and $\recgap(M,T)$, and we
want them for values of $T$ that reach the advertised window. The claim
under test is that the mechanisms in Kimi K3 drive $\sinkmass$ toward zero
and $\recgap$ toward zero together. Our position is that these are two
claims, not one, and that they need two measurements.

% ============================================
% 4. METHODOLOGY
% ============================================
\section{Methodology}

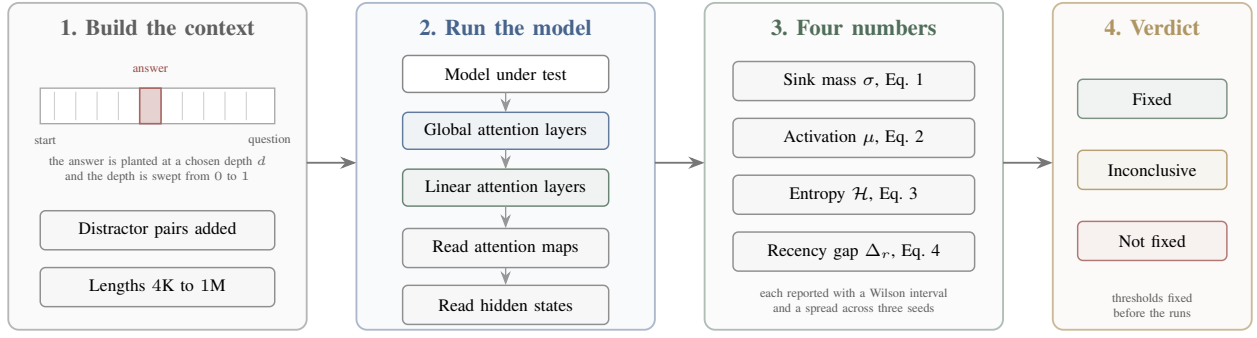
\begin{figure*}[!t]
\centering
\begin{tikzpicture}[scale=0.94, transform shape]

% ---------- stage 1. build the context ----------
\draw[rounded corners=4pt, draw=inkColor!45, fill=inkColor!3, line width=0.7pt]
  (0.10,-1.05) rectangle (4.30,3.55);
\node[font=\small\bfseries, text=inkColor] at (2.20,3.20) {1. Build the context};

\draw[draw=inkColor!55, fill=white, line width=0.5pt]
  (0.55,1.85) rectangle (3.85,2.35);
\foreach \x in {0.75,1.05,...,3.65} {
  \draw[line width=0.4pt, inkColor!30] (\x,1.90) -- (\x,2.30);
}
\draw[draw=sinkColor!85, fill=sinkColor!25, line width=0.6pt]
  (1.95,1.85) rectangle (2.25,2.35);
\node[tinytag, text=sinkColor] at (2.10,2.62) {answer};
\node[tinytag] at (0.62,1.62) {start};
\node[tinytag] at (3.78,1.62) {question};
\node[tinytag, align=center] at (2.20,1.20)
  {the answer is planted at a chosen depth $d$\\and the depth is swept
   from $0$ to $1$};
\node[probe, minimum width=3.3cm] at (2.20,0.35) {Distractor pairs added};
\node[probe, minimum width=3.3cm] at (2.20,-0.45)
  {Lengths $4$K to $1$M};

% ---------- stage 2. run the model ----------
\draw[rounded corners=4pt, draw=gateColor!45, fill=gateColor!3, line width=0.7pt]
  (5.00,-1.05) rectangle (9.20,3.55);
\node[font=\small\bfseries, text=gateColor] at (7.10,3.20) {2. Run the model};

\node[layerbox, minimum width=2.9cm] at (7.10,2.55) {Model under test};
\node[mlabox, minimum width=2.9cm] at (7.10,1.75) {Global attention layers};
\node[kdabox, minimum width=2.9cm] at (7.10,0.95) {Linear attention layers};
\node[probe, minimum width=2.9cm] at (7.10,0.10) {Read attention maps};
\node[probe, minimum width=2.9cm] at (7.10,-0.70) {Read hidden states};
\draw[thinflow] (7.10,2.29) -- (7.10,2.01);
\draw[thinflow] (7.10,1.49) -- (7.10,1.21);
\draw[thinflow] (7.10,0.69) -- (7.10,0.38);
\draw[thinflow] (7.10,-0.18) -- (7.10,-0.42);

% ---------- stage 3. four numbers ----------
\draw[rounded corners=4pt, draw=kdaColor!45, fill=kdaColor!3, line width=0.7pt]
  (9.90,-1.05) rectangle (14.10,3.55);
\node[font=\small\bfseries, text=kdaColor] at (12.00,3.20) {3. Four numbers};

\node[probe, minimum width=3.4cm] at (12.00,2.45)
  {Sink mass $\sinkmass$, Eq.~\ref{eq:sink}};
\node[probe, minimum width=3.4cm] at (12.00,1.65)
  {Activation $\maxact$, Eq.~\ref{eq:act}};
\node[probe, minimum width=3.4cm] at (12.00,0.85)
  {Entropy $\mathcal{H}$, Eq.~\ref{eq:entropy}};
\node[probe, minimum width=3.4cm] at (12.00,0.05)
  {Recency gap $\recgap$, Eq.~\ref{eq:recency}};
\node[tinytag, align=center] at (12.00,-0.65)
  {each reported with a Wilson interval\\and a spread across three seeds};

% ---------- stage 4. verdict ----------
\draw[rounded corners=4pt, draw=warnColor!55, fill=warnColor!4, line width=0.7pt]
  (14.80,-1.05) rectangle (17.60,3.55);
\node[font=\small\bfseries, text=warnColor!85, align=center]
  at (16.20,3.20) {4. Verdict};

\node[probe, minimum width=2.1cm, draw=kdaColor!75, fill=kdaColor!8]
  at (16.20,2.20) {Fixed};
\node[probe, minimum width=2.1cm, draw=warnColor!75, fill=warnColor!8]
  at (16.20,1.20) {Inconclusive};
\node[probe, minimum width=2.1cm, draw=sinkColor!75, fill=sinkColor!8]
  at (16.20,0.20) {Not fixed};
\node[tinytag, align=center] at (16.20,-0.70)
  {thresholds fixed\\before the runs};

% ---------- connectors ----------
\draw[flow] (4.30,1.30) -- (5.00,1.30);
\draw[flow] (9.20,1.30) -- (9.90,1.30);
\draw[flow] (14.10,1.30) -- (14.80,1.30);

\end{tikzpicture}
\caption{The SinkProbe pipeline. A context is built with the answer
planted at a known depth, the model reads it while the suite records
attention maps and hidden states, four numbers are computed from those
recordings, and each number is compared against a threshold that was
fixed before any large run took place. The same four stages apply
unchanged to the small models we train ourselves and to a released
checkpoint, which is what makes the pilot and the full scale protocol
comparable.}
\label{fig:pipeline}
\end{figure*}

SinkProbe is the diagnostic suite that produces the four numbers, and the
model ladder is the controlled setting in which we validate it.
Figure~\ref{fig:pipeline} shows how the pieces fit together.

% ============================================
% FIGURE. SINKPROBE PIPELINE
% ============================================

\subsection{Design Goals}

Three goals shaped the design, and each one ruled something out.

\begin{enumerate}[leftmargin=*]
  \item \textbf{One change at a time.} Any difference between two
  measurements should be attributable to one architectural change. This
  ruled out comparing released checkpoints against each other, since they
  differ in data, scale and training budget as well as in mechanism.
  \item \textbf{Runnable without a cluster.} The pilot has to be
  reproducible by a reader with a laptop, otherwise nobody checks it. This
  ruled out any design that needs pre-trained weights or a GPU.
  \item \textbf{Decided in advance.} The thresholds that separate a pass
  from a failure are fixed before the large runs. This ruled out reporting
  the large scale numbers first and interpreting them afterwards.
\end{enumerate}

\subsection{The Model Ladder}

We train four models that differ only in how they mix tokens and depth.
Width, depth, head count, feed forward size, vocabulary, data, optimiser,
schedule and seed policy are identical across all four.

\begin{itemize}[leftmargin=*]
  \item \textbf{Softmax.} Eight layers of causal softmax attention with
  rotary position embeddings \cite{su2024roformer}. This is the baseline
  that should form a sink.
  \item \textbf{Softmax with gate.} The same stack with a channel wise
  input dependent sigmoid gate on the attention output, which is the
  mechanism of \cite{qiu2025gated}.
  \item \textbf{Hybrid.} Three linear delta style layers followed by one
  gated global layer, repeated twice, with no position encoding. This is
  the Kimi K3 token mixing pattern at small scale \cite{kimi2026k3}.
  \item \textbf{Hybrid with attention residuals.} The same stack with
  Block Attention Residuals, given in Equation~\ref{eq:attnres},
  replacing the running sum over depth \cite{kimi2026attnres}. Tables and
  figures shorten this to AttnRes.
\end{itemize}

Each step of the ladder adds exactly one mechanism, so the differences
between neighbouring rows in the results tables are ablations by
construction rather than by a separate experiment.

\subsection{The Linear Layer We Use}

Our linear layer keeps a state $S_t \in \R^{d_k \times d_v}$ and updates
it with a channel wise decay,
\begin{equation}
S_t \;=\; \mathrm{Diag}(a_t)\, S_{t-1} \;+\; k_t v_t^{\top},
\qquad o_t \;=\; \bigl(S_t^{\top} q_t\bigr) \odot g_t ,
\label{eq:kda}
\end{equation}
where $a_t \in (0,1]^{d_k}$ comes from the input and $g_t$ is a full rank
output gate. We bound $\log a_t$ from below so that the chunked form stays
stable in single precision, which follows the lower bounded decay used in
Kimi K3.

How the decay starts turned out to matter, and we record it because it
cost us two runs to find. The decay controls two abilities at once. A
channel that fades quickly records how long ago something happened, and a
channel that barely fades carries content across a long span. Starting
every channel at the fast end leaves the layer with order and no memory.
Starting every channel at the slow end leaves it with memory and no
order. We therefore spread the channels across the range at
initialisation, from a decay near $0.78$ to one near $0.999$, and let
training move them.

The layer is a simplification and we state it plainly. The real Kimi
Delta Attention layer also applies a delta rule correction that removes
the stale part of the state before writing to it \cite{kimi2025linear}.
We leave that term out, so our pilot measures what a decaying fixed size
state does, not what the exact Kimi kernel does. Section VIII reports what
that costs, and the cost is larger than we expected.

\subsection{Attention Residuals Over Depth}

For Block Attention Residuals we follow \cite{kimi2026attnres} directly.
The eight layers are split into two blocks of four. Each block is reduced
to one representation $b_n$ by summing its layer outputs, and the token
embedding is kept as $b_0$ so that the input is always reachable. Layer
$\ell$ then builds its own input by attending over the sources available
to it,
\begin{equation}
\alpha^{\ell}_{i} \;=\;
\frac{\exp\!\bigl(w_\ell^{\top}\,\mathrm{RMSNorm}(b_i)\bigr)}
     {\sum_j \exp\!\bigl(w_\ell^{\top}\,\mathrm{RMSNorm}(b_j)\bigr)},
\qquad
h_\ell \;=\; \sum_i \alpha^{\ell}_i\, b_i ,
\label{eq:attnres}
\end{equation}
with $w_\ell$ a learned pseudo query for that layer and RMSNorm (root
mean square normalisation) applied to each source. That normalisation
stops a block with large magnitude outputs from winning the weights on
magnitude alone.

\subsection{The Retrieval Probe}

Each test sequence is a haystack of filler tokens with one key and value
pair planted at a chosen depth, several distractor pairs scattered
elsewhere, and a question at the end that names the key. The model has to
return the matching value. Depth is controlled directly, so $p(d)$ is
measured rather than inferred.

Three design choices keep the probe honest. Distractor pairs use
different keys, so a model cannot answer by copying the only pair it saw.
The answer is scored at one position only, so partial credit for fluent
filler is impossible. And we report two floors with every result, blind
chance and the score a model would get by returning any value present in
the context, because clearing the first floor is easy and clearing the
second is what retrieval means.

The pilot also runs one control that changes the objective rather than
the architecture. It trains the baseline stack with the auxiliary next
token term removed, leaving only the answer position in the loss. If the
sink is created by the pressure to produce an output at every position,
that control should show no sink at all while learning the same task.
Section VI reports the outcome.

\subsection{Statistical Treatment}

Every accuracy is a binomial proportion, so we report Wilson score
intervals rather than a plain standard error \cite{wilson1927}. For a
target half width $\epsilon$ at the worst case proportion, the number of
independent trials needed is
\begin{equation}
n \;\ge\; \frac{z^2\,p(1-p)}{\epsilon^{2}}
\;\approx\; \frac{0.96}{\epsilon^{2}} \quad
\text{at } p = 0.5,\; z = 1.96 .
\label{eq:power}
\end{equation}
Table~\ref{tab:power} turns this into the trial counts used later. Every
result is run with three seeds, and we report the spread across seeds
next to the interval within a seed, because the two sources of variation
answer different questions.

\begin{table}[!t]
\centering
\caption{Trials needed for a given confidence half width, from
Equation~\ref{eq:power} at the worst case proportion. The sweep uses
eleven depth points, and the last column is the cost of the one million
token row alone. We operate at seven points, marked in bold, because the
cost of three points is more than five times higher for a gain that no
threshold in Table~\ref{tab:thresholds} depends on.}
\label{tab:power}
\footnotesize
\renewcommand{\arraystretch}{1.15}
\begin{tabular}{@{}cccc@{}}
\toprule
\textbf{Half} & \textbf{Trials per} & \textbf{Sequences} &
\textbf{Tokens read} \\
\textbf{width} & \textbf{depth} & \textbf{in total} &
\textbf{at $T = 1$M} \\
\midrule
$\pm 10$ points & 97 & 1\,067 & 1.07 billion \\
$\pm 7$ points & \textbf{196} & \textbf{2\,156} & \textbf{2.16 billion} \\
$\pm 5$ points & 385 & 4\,235 & 4.24 billion \\
$\pm 3$ points & 1\,068 & 11\,748 & 11.75 billion \\
$\pm 2$ points & 2\,401 & 26\,411 & 26.41 billion \\
\bottomrule
\end{tabular}
\end{table}

\subsection{Thresholds Registered in Advance}

Table~\ref{tab:thresholds} states what will count as a pass and what will
count as a failure when the suite is applied to the released weights. The
numbers come from the published values in \cite{qiu2025gated}, so they are
not tuned to anything we measured. Fixing them now is the point.

\begin{table}[!t]
\centering
\caption{Thresholds fixed before the full scale runs. A result between
the two columns is reported as inconclusive rather than argued either
way.}
\label{tab:thresholds}
\footnotesize
\renewcommand{\arraystretch}{1.15}
\begin{tabular}{@{}>{\raggedright\arraybackslash}p{2.5cm}>{\raggedright\arraybackslash}p{2.0cm}>{\raggedright\arraybackslash}p{2.4cm}@{}}
\toprule
\textbf{Quantity} & \textbf{Counts as fixed} & \textbf{Counts as not
fixed} \\
\midrule
Sink mass $\sinkmass$ & below $0.05$ at every length & above $0.20$ at any
length \\
Worst layer sink & below $0.15$ & above $0.40$ \\
Massive activation $\maxact$ & below $150$ & above $600$ \\
Recency gap $\recgap$ & within $\pm 5$ points & beyond $\pm 15$ points \\
Recall at mid depth & within 10 points of end depth & more than 25 points
below \\
\bottomrule
\end{tabular}
\end{table}

% ============================================
% 5. EXPERIMENTAL SETUP
% ============================================
\section{Experimental Setup}

\subsection{Models and Training}

All four models share the configuration in Table~\ref{tab:setup}. They
are small on purpose. The question they answer is whether a mechanism
changes a diagnostic, not what score a frontier model reaches, and a
small model that a reader can retrain in minutes is worth more here than
a large one they have to take on trust.

\begin{table}[!t]
\centering
\caption{Configuration shared by all four models. Only the mixing rule
differs between them.}
\label{tab:setup}
\footnotesize
\renewcommand{\arraystretch}{1.15}
\begin{tabular}{@{}l>{\raggedright\arraybackslash}p{4.3cm}@{}}
\toprule
\textbf{Setting} & \textbf{Value} \\
\midrule
Layers & 8 \\
Model width & 96 \\
Attention heads & 4, head width 24 \\
Feed forward width & 256, SwiGLU (swish gated linear unit) \\
Normalisation & RMSNorm (root mean square), pre norm \\
Vocabulary & 70 tokens \\
Blocks for attention residuals & 2 blocks of 4 layers \\
Chunk size for linear layers & 32 \\
Training length & 96 tokens \\
Evaluation lengths & 96, 192, 384, 768 \\
Optimiser & AdamW (Adam with decoupled weight decay) \\
Optimiser settings & $\beta = (0.9, 0.95)$, decay 0.01 \\
Learning rate & $2\times 10^{-3}$, one cycle, 8 percent warmup \\
Gradient clipping & 1.0 \\
Batch size & 16 \\
Training steps & 1500 \\
Haystack motif and noise & period 24, half the positions random \\
Auxiliary next token weight & 0.3, and 0.0 for the control \\
Seeds & 3 per model \\
Precision & float32 \\
Hardware & 16 thread laptop processor, no accelerator \\
\bottomrule
\end{tabular}
\end{table}

Three settings deserve a note.

The auxiliary next token objective matters more than its small weight
suggests. A sink forms when many query positions have to produce an
output while having nothing worth reading, and a task scored at one
position alone never creates those positions. We therefore add an
ordinary next token term across the whole sequence. To check that this is
the mechanism rather than an assumption, we also train a control with the
auxiliary weight set to zero and everything else identical. Section VI
reports what that control does to the sink.

The haystack is partly predictable rather than uniform. Each sequence
repeats its own random motif of 24 symbols, and half the positions are
then replaced at random. A uniform haystack would make the auxiliary
term unlearnable, so it would inject noise instead of the partial
predictability that ordinary text has. With a motif, prediction inside
the haystack is solvable by looking back one period, which is the same
kind of lookup the question needs.

How predictable the haystack is turned out to matter more than we
expected, and we record it here because it shaped the setting. In a
single seed calibration run, a haystack with only 15 percent of positions
randomised produced a sink mass of 0.095 in the baseline, while the same
model on a haystack with half the positions randomised produced 0.438.
The reading is straightforward and it fits the account in Section III. A
head only needs somewhere to dump its budget when there is nothing worth
reading, so the sink grows with the share of the context that is not
worth reading. We fixed the noise at one half because the resulting
baseline sits closest to the value reported for real models on real text
\cite{qiu2025gated}. This was a single seed observation used to pick a
setting, not a result, and we do not report it as one.

The evaluation lengths run to eight times the training length so that
extrapolation is visible. That is the small scale analogue of a model
trained at 64 thousand tokens being asked to work at one million.

\subsection{The Retrieval Task}

Sequences use 12 distinct keys, 12 distinct values, 3 distractor pairs
and 44 filler symbols. Two floors matter when reading any accuracy in
this paper, and we give both with every table. Blind chance is $1/12$,
which is 8.3 percent. A model that ignores the question and returns any
value it saw in the context scores $1/4$, which is 25 percent, because
four pairs are present. A result is only evidence of retrieval if it
clears the second floor, not merely the first.

The answer depth is drawn uniformly for the aggregate numbers and fixed
on a grid of ten points for the depth profiles.

\subsection{Evaluation Protocol}

Each model is evaluated at four lengths with 128 sequences per length per
seed, giving 384 sequences per length once the three seeds are pooled. A
Wilson interval at that count has a half width of about 5 points at the
worst case proportion. Depth profiles use 32 sequences per depth point
per seed, which is deliberately coarser, because the profile is read for
its shape rather than for the value at any single point.

The diagnostics in Equations~\ref{eq:sink} to \ref{eq:entropy} are
recorded on the first evaluation batch at each length, since they are
averages over heads, query positions and layers, and are already tightly
determined by 16 sequences of a few hundred tokens each.

\subsection{Cache Growth for the Kimi K3 Layer Mix}

The cost side of the claim can be checked without running the model at
all, from the layer counts in the technical report \cite{kimi2026k3}. Of
93 layers, 69 are linear and 24 are global. A linear layer holds a state
whose size depends on the head geometry and not on the context, and a
global layer holds a cache that grows in proportion to the context. So
74.2 percent of the stack contributes nothing to cache growth, and that
figure needs no assumption at all.

Turning the fraction into bytes needs two numbers the report does not
publish, namely the latent rank of the global layers and the head
geometry of the state. We take a latent rank of 512 and a state of
$128 \times 128$ per head across 96 heads, both stated as assumptions and
both easy to substitute later. Table~\ref{tab:cache} reports the result
in bfloat16, alongside what a fully dense stack of the same depth and
width would need.

\begin{table}[!t]
\centering
\caption{Cache growth for the Kimi K3 layer mix. The state column is flat
because 69 of the 93 layers hold a cache whose size does not depend on
context length. The dense column is what the same depth would cost with
ordinary key and value caching. Values assume bfloat16, a latent rank of
512 and a $128 \times 128$ state per head.}
\label{tab:cache}
\footnotesize
\renewcommand{\arraystretch}{1.15}
\begin{tabular}{@{}rrrrr@{}}
\toprule
\textbf{Context} & \textbf{Global} & \textbf{Linear} & \textbf{Hybrid} &
\textbf{Dense} \\
\textbf{tokens} & \textbf{cache} & \textbf{state} & \textbf{total} &
\textbf{stack} \\
 & (GiB) & (GiB) & (GiB) & (GiB) \\
\midrule
4\,096 & 0.09 & 0.20 & 0.30 & 10.1 \\
32\,768 & 0.75 & 0.20 & 0.95 & 80.7 \\
131\,072 & 3.00 & 0.20 & 3.20 & 322.6 \\
262\,144 & 6.00 & 0.20 & 6.20 & 645.2 \\
524\,288 & 12.00 & 0.20 & 12.20 & 1\,290.4 \\
1\,048\,576 & 24.00 & 0.20 & 24.20 & 2\,580.8 \\
\bottomrule
\end{tabular}
\end{table}

The reading is straightforward. At one million tokens the hybrid stack
needs about 24 GiB (gibibytes) of cache where a dense stack of the same
depth would
need about 2.5 TiB, a factor of roughly 107. Below about 8 thousand
tokens the fixed state is the larger of the two terms, so the design only
pays for itself once the context is long, which is exactly the regime it
was built for.

\subsection{What This Setup Cannot Show}

The pilot runs at 768 tokens, not at one million, and its models hold
about one million parameters rather than 2.8 trillion. It can show that a
mechanism moves a diagnostic in a controlled comparison, and it can show
when a mechanism fails to move one. It cannot show the size of either at
frontier scale, and we do not extrapolate. Section VII states what has to
be run to answer that part, and Section VIII lists the ways the pilot
could mislead.

% ============================================
% 6. RESULTS
% ============================================
\section{Results}

We report five things in order. Whether the sink appears at all and what
causes it. What each mechanism does to it. What happens to recall as the
context grows past the training length. Where in the context the evidence
has to sit for the model to find it. And what the whole thing costs.

\subsection{The Sink is Made by the Objective}

Before asking what an architecture does to a sink, it is worth asking
what puts one there. Our control changes the objective and nothing else.
The same eight layer softmax stack reads the same data with the same
three seeds, and the only difference is whether the loss covers every
position in the sequence or only the position that holds the answer.

Table~\ref{tab:control} gives the result and it is not a small effect.
Training the model to predict at every position leaves 31.6 percent of
attention on the first token, with the worst layer at 63.3 percent.
Training it to answer only the final question leaves 4.9 percent, with
the worst layer at 6.7 percent. That is a factor of six on the average
and a factor of nine on the worst layer, from a change that touches no
weight shape and no layer.

Two details make the comparison stronger than the headline. Recall is
unchanged, at 33.3 percent without the auxiliary term against 32.0
percent with it, so the sink was not paying for any of the retrieval the
model actually does. And the spread across seeds collapses. With the
auxiliary term the three seeds landed at 0.455, 0.089 and 0.405. Without
it they landed at 0.051, 0.046 and 0.050. The sink is not only smaller,
it stops being a coin toss.

This follows from the definition in Section~III. A softmax head has to
spend a budget of one at every position it is asked to produce an output
for. Take most of those positions out of the loss and most of the
pressure goes with them.

\begin{table}[!t]
\centering
\caption{The control that changes the objective instead of the architecture. Both rows use the same softmax stack, the same data and the same seeds. The only difference is whether the model has to predict at every position or only at the answer. Values at the training length.}
\label{tab:control}
\footnotesize
\renewcommand{\arraystretch}{1.15}
\begin{tabular}{@{}lcccc@{}}
\toprule
\textbf{Objective} & \textbf{Sink} & \textbf{Worst} & \textbf{Activation} & \textbf{Recall} \\
 & mass & layer & $\maxact$ & (\%) \\
\midrule
Answer position only & 0.049 $\pm$ 0.003 & 0.067 & 26.2 & 33.3 \\
Every position & 0.316 $\pm$ 0.198 & 0.633 & 74.9 & 32.0 \\
\bottomrule
\end{tabular}
\end{table}

\subsection{Every Diagnostic, Every Model, Every Length}

Table~\ref{tab:main} carries every diagnostic for the four models at
every evaluation length. Three things in it need saying plainly before
any comparison is drawn.

The baseline does form a substantial sink. Averaged over three seeds and
its eight softmax layers, our untreated stack puts 31.6 percent of its
attention on the first position at the training length, and its worst
layer puts 63.3 percent. A stack reading its context evenly would put
about 1.0 percent there, so this is roughly thirty times an even split.
It is also the same order of magnitude as the 46.7 percent average and 83
percent worst layer that \cite{qiu2025gated} measured in a fifteen
billion parameter model, which is worth noting given that our models are
four orders of magnitude smaller.

The spread across seeds is very large. Individual baseline runs landed at
sink masses of 0.455, 0.089 and 0.405. The same architecture on the same
data with the same schedule produced one run with almost no sink and two
with a large one. Any comparison between architectures has to clear that
spread before it means anything, and in this table most of them do not.

The hybrid rows did not learn the task. Their recall sits at 6.0 and 5.7
percent against a blind chance level of 8.3 percent, so they are not
retrieving at all, let alone clearing the 25 percent floor that returning
any value present in the context would give. Section~VIII explains why,
and the short version is that we left out the delta rule correction that
makes linear attention competitive at this kind of recall. Their sink,
activation and entropy figures are still measurements of a trained stack
and are reported as such. Their recall and recency figures describe a
model that never solved the probe and carry no information about position
bias.

\begin{table*}[!t]
\centering
\caption{Every diagnostic for the four models at four evaluation lengths. Recall is pooled over 3 seeds and shown with the spread across seeds. Blind chance is 8.3 percent and returning any value present in the context scores 25.0 percent, so only the distance above the second floor is retrieval. Sink mass averages the softmax layers only, and the count of those layers is given in the last column.}
\label{tab:main}
\footnotesize
\renewcommand{\arraystretch}{1.15}
\begin{tabular}{@{}llccccccc@{}}
\toprule
\textbf{Model} & \textbf{Length} & \textbf{Recall} & \textbf{Sink mass} & \textbf{Worst layer} & \textbf{Activation} & \textbf{Entropy} & \textbf{Recency} & \textbf{Softmax} \\
 & (tokens) & (\%) & $\sinkmass$ & sink & $\maxact$ & $\mathcal{H}$ & gap & layers \\
\midrule
\textbf{Softmax} & 96 & 32.0 $\pm$ 3.9 & 0.316 & 0.633 & 74.9 & 1.63 & -10.9 & 8 \\
 & 192 & 27.1 $\pm$ 2.4 & 0.239 & 0.511 & 74.5 & 2.33 & +6.4 & 8 \\
 & 384 & 34.9 $\pm$ 3.0 & 0.183 & 0.415 & 74.2 & 3.02 & +1.0 & 8 \\
 & 768 & 26.3 $\pm$ 8.6 & 0.126 & 0.301 & 73.4 & 3.76 & +15.2 & 8 \\
\midrule
\textbf{Softmax + gate} & 96 & 28.1 $\pm$ 2.1 & 0.261 & 0.490 & 36.2 & 1.40 & -15.9 & 8 \\
 & 192 & 29.7 $\pm$ 5.1 & 0.211 & 0.381 & 35.9 & 2.08 & -6.6 & 8 \\
 & 384 & 30.5 $\pm$ 3.1 & 0.160 & 0.311 & 34.7 & 2.74 & +10.3 & 8 \\
 & 768 & 31.8 $\pm$ 3.7 & 0.103 & 0.231 & 34.0 & 3.39 & +0.2 & 8 \\
\midrule
\textbf{Hybrid 3:1} & 96 & 6.0 $\pm$ 2.7 & 0.183 & 0.233 & 97.9 & 2.25 & +1.5 & 2 \\
 & 192 & 8.1 $\pm$ 1.6 & 0.132 & 0.171 & 97.9 & 3.05 & -3.4 & 2 \\
 & 384 & 9.9 $\pm$ 3.2 & 0.101 & 0.134 & 97.9 & 3.67 & +6.1 & 2 \\
 & 768 & 7.3 $\pm$ 3.2 & 0.075 & 0.107 & 97.7 & 4.26 & -3.2 & 2 \\
\midrule
\textbf{Hybrid + AttnRes} & 96 & 5.7 $\pm$ 0.9 & 0.202 & 0.251 & 38.9 & 2.48 & -1.8 & 2 \\
 & 192 & 8.9 $\pm$ 0.5 & 0.131 & 0.159 & 38.9 & 3.14 & -6.9 & 2 \\
 & 384 & 9.1 $\pm$ 2.4 & 0.079 & 0.091 & 39.0 & 3.78 & +0.6 & 2 \\
 & 768 & 7.0 $\pm$ 0.8 & 0.046 & 0.053 & 38.9 & 4.38 & -5.5 & 2 \\
\bottomrule
\end{tabular}
\end{table*}

\subsection{What Each Mechanism Contributes}

\begin{figure}[!t]
\centering
\begin{tikzpicture}
\begin{axis}[
    width=0.99\columnwidth,
    height=4.2cm,
    xlabel={Evaluation length in tokens},
    ylabel={Sink mass $\sinkmass$},
    xlabel style={font=\scriptsize},
    ylabel style={font=\scriptsize},
    xmode=log, log basis x=2,
    xtick={96,192,384,768},
    xticklabels={96,192,384,768},
    x tick label style={font=\tiny},
    y tick label style={font=\tiny},
    ymin=0,
    legend style={at={(0.98,0.97)}, anchor=north east, font=\tiny,
                  draw=inkColor!40, fill=white, row sep=0pt},
    legend cell align=left,
    grid=major, grid style={gray!20},
    axis line style={inkColor!70},
]
\addplot[color=sinkColor, mark=*, mark size=1.5pt, line width=1.0pt] coordinates {(96,0.3162) (192,0.2386) (384,0.1827) (768,0.1264)};
\addplot[color=gateColor, mark=square*, mark size=1.5pt, line width=1.0pt] coordinates {(96,0.2609) (192,0.2106) (384,0.1600) (768,0.1034)};
\addplot[color=kdaColor, mark=triangle*, mark size=1.5pt, line width=1.0pt] coordinates {(96,0.1831) (192,0.1317) (384,0.1014) (768,0.0754)};
\addplot[color=depthColor, mark=diamond*, mark size=1.5pt, line width=1.0pt] coordinates {(96,0.2016) (192,0.1315) (384,0.0791) (768,0.0463)};
\legend{Softmax, Softmax + gate, Hybrid 3:1, Hybrid + AttnRes}
\end{axis}
\end{tikzpicture}
\caption{Sink mass against evaluation length. A stack that read its
context evenly would sit at one over the length, which is 0.010 at 96
tokens and 0.001 at 768, so every curve above the axis is spending more
on the first position than an even split would. Sink mass averages
softmax layers only, so the hybrid curves average two layers where the
dense curves average eight.}
\label{fig:sinklen}
\end{figure}
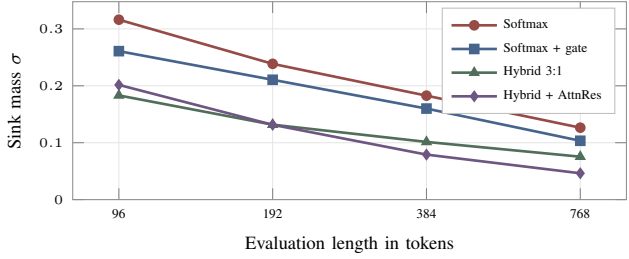

Table~\ref{tab:ablation} walks up the ladder one mechanism at a time and
Figure~\ref{fig:sinklen} plots sink mass against evaluation length.

Sink mass does fall as mechanisms are added, from 0.316 for the plain
softmax stack to 0.261 with the gate and 0.183 for the hybrid. The worst
layer falls further and more cleanly, from 0.633 to 0.490 to 0.233. But
the baseline seeds ranged from 0.089 to 0.455, so a drop of 0.055 from
adding the gate is a fraction of the noise it sits in, and the gated run
with the highest sink scored above two of the three baseline runs. We
report the gate comparison as inconclusive. The hybrid comparison is
larger and cleaner on the worst layer, and it is also confounded, because
the hybrid never learned the task and its heads therefore had different
work to do.

The gate result needs an explanation, because at fifteen billion
parameters the same mechanism cuts first token attention by more than a
factor of nine \cite{qiu2025gated}. The suite gives one in a number
rather than a guess. The gate works by learning to be sparse, so that a
head can emit nothing when it has nothing to say. The mean gate score
reported in \cite{qiu2025gated} is 0.116 after trillions of tokens, a
gate closed most of the time. Ours sits at 0.686 after fifteen hundred
steps, a gate open most of the time. The mechanism was in the
architecture and absent from the trained weights, so every head kept
emitting at every position and the condition that creates a sink was
never removed.

This is the argument against answering the question in the title with a
stand in. If a mechanism known to remove sinks does not remove them in a
small model trained briefly, no small model can speak for the one we want
to measure.

\begin{table}[!t]
\centering
\caption{Each row adds one mechanism to the row above it. The third column compares sink mass against the untreated baseline in the first row. Values at the training length of 96 tokens.}
\label{tab:ablation}
\footnotesize
\renewcommand{\arraystretch}{1.15}
\begin{tabular}{@{}lccc@{}}
\toprule
\textbf{Model} & \textbf{Sink mass} & \textbf{Against} & \textbf{Recency gap} \\
 & $\sinkmass$ & baseline & (points) \\
\midrule
Softmax & 0.316 $\pm$ 0.198 & baseline & -10.9 $\pm$ 16.5 \\
Softmax + gate & 0.261 $\pm$ 0.175 & -18\% & -15.9 $\pm$ 13.4 \\
Hybrid 3:1 & 0.183 $\pm$ 0.103 & -42\% & +1.5 $\pm$ 2.3 \\
Hybrid + AttnRes & 0.202 $\pm$ 0.180 & -36\% & -1.8 $\pm$ 5.3 \\
\bottomrule
\end{tabular}
\end{table}

\subsection{Recall Past the Training Length}

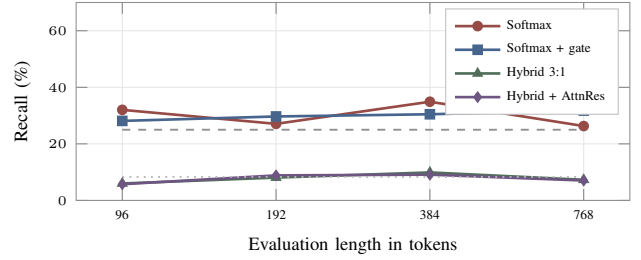
\begin{figure}[!t]
\centering
\begin{tikzpicture}
\begin{axis}[
    width=0.99\columnwidth,
    height=4.2cm,
    xlabel={Evaluation length in tokens},
    ylabel={Recall (\%)},
    xlabel style={font=\scriptsize},
    ylabel style={font=\scriptsize},
    xmode=log, log basis x=2,
    xtick={96,192,384,768},
    xticklabels={96,192,384,768},
    x tick label style={font=\tiny},
    y tick label style={font=\tiny},
    ymin=0, ymax=70,
    legend style={at={(0.98,0.97)}, anchor=north east, font=\tiny,
                  draw=inkColor!40, fill=white, row sep=0pt},
    legend cell align=left,
    grid=major, grid style={gray!20},
    axis line style={inkColor!70},
]
\addplot[color=sinkColor, mark=*, mark size=1.5pt, line width=1.0pt] coordinates {(96,32.03) (192,27.08) (384,34.90) (768,26.30)};
\addplot[color=gateColor, mark=square*, mark size=1.5pt, line width=1.0pt] coordinates {(96,28.12) (192,29.69) (384,30.47) (768,31.77)};
\addplot[color=kdaColor, mark=triangle*, mark size=1.5pt, line width=1.0pt] coordinates {(96,5.99) (192,8.07) (384,9.90) (768,7.29)};
\addplot[color=depthColor, mark=diamond*, mark size=1.5pt, line width=1.0pt] coordinates {(96,5.73) (192,8.85) (384,9.11) (768,7.03)};
\addplot[color=inkColor!70, dashed, line width=0.7pt, forget plot]
  coordinates {(96,25.0) (768,25.0)};
\addplot[color=inkColor!45, dotted, line width=0.7pt, forget plot]
  coordinates {(96,8.3) (768,8.3)};
\legend{Softmax, Softmax + gate, Hybrid 3:1, Hybrid + AttnRes}
\end{axis}
\end{tikzpicture}
\caption{Recall against evaluation length. Models are trained at 96
tokens and tested out to 768, which is eight times further. The dashed
line at 25 percent is the score for returning any value present in the
context and the dotted line at 8.3 percent is blind chance. Only the
distance above the dashed line counts as retrieval.}
\label{fig:acclen}
\end{figure}

Figure~\ref{fig:acclen} plots recall against evaluation length. The models
are trained at 96 tokens and tested out to 768, which is eight times
further, and the two horizontal guides mark the floors that matter. Only
distance above the upper guide counts as retrieval.

The two softmax stacks stay above that guide across the range, at 32.0
and 28.1 percent at the training length and 26.3 and 31.8 percent at 768,
so the retrieval behaviour they learned is not tied to the length they
saw. The baseline at 768 carries a seed spread of 8.6 points and sits
close enough to the 25 percent floor that we would not claim retrieval
survives at that length for that model. The gated stack, which is the
weaker of the two at the training length, is the stronger at 768.

Sink mass falls steadily across the same range in
Figure~\ref{fig:sinklen}, and that fall is mostly arithmetic. Attention
on the first position is a share of a budget spread over more positions,
so an unchanged absolute preference reads as a smaller fraction at a
longer length. The ratio against an even split is the quantity that stays
interpretable, and it moves the other way. At 96 tokens the baseline puts
about thirty times an even share on the first position, and at 768 tokens
it puts about ninety seven times an even share there. Reported as a
fraction the sink looks like it is going away with length. Reported
against what an even reader would do, it is getting worse.

\subsection{Where the Evidence Has to Sit}

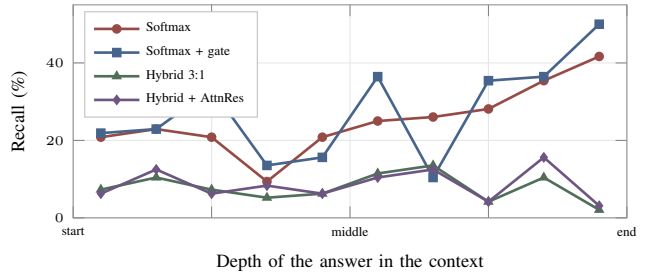
\begin{figure}[!t]
\centering
\begin{tikzpicture}
\begin{axis}[
    width=0.99\columnwidth,
    height=4.4cm,
    xlabel={Depth of the answer in the context},
    ylabel={Recall (\%)},
    xlabel style={font=\scriptsize},
    ylabel style={font=\scriptsize},
    xmin=0, xmax=1,
    xtick={0,0.25,0.5,0.75,1.0},
    xticklabels={start,,middle,,end},
    x tick label style={font=\tiny},
    y tick label style={font=\tiny},
    ymin=0,
    legend style={at={(0.02,0.97)}, anchor=north west, font=\tiny,
                  draw=inkColor!40, fill=white, row sep=0pt},
    legend cell align=left,
    grid=major, grid style={gray!20},
    axis line style={inkColor!70},
]
\addplot[color=sinkColor, mark=*, mark size=1.3pt, line width=1.0pt] coordinates {(0.05,20.83) (0.15,22.92) (0.25,20.83) (0.35,9.38) (0.45,20.83) (0.55,25.00) (0.65,26.04) (0.75,28.12) (0.85,35.42) (0.95,41.67)};
\addplot[color=gateColor, mark=square*, mark size=1.3pt, line width=1.0pt] coordinates {(0.05,21.88) (0.15,22.92) (0.25,33.33) (0.35,13.54) (0.45,15.62) (0.55,36.46) (0.65,10.42) (0.75,35.42) (0.85,36.46) (0.95,50.00)};
\addplot[color=kdaColor, mark=triangle*, mark size=1.3pt, line width=1.0pt] coordinates {(0.05,7.29) (0.15,10.42) (0.25,7.29) (0.35,5.21) (0.45,6.25) (0.55,11.46) (0.65,13.54) (0.75,4.17) (0.85,10.42) (0.95,2.08)};
\addplot[color=depthColor, mark=diamond*, mark size=1.3pt, line width=1.0pt] coordinates {(0.05,6.25) (0.15,12.50) (0.25,6.25) (0.35,8.33) (0.45,6.25) (0.55,10.42) (0.65,12.50) (0.75,4.17) (0.85,15.62) (0.95,3.12)};
\legend{Softmax, Softmax + gate, Hybrid 3:1, Hybrid + AttnRes}
\end{axis}
\end{tikzpicture}
\caption{Recall against the position of the answer inside the context, at
the longest evaluation length. A flat line means the whole window is read
evenly. A line that rises to the right means the model prefers evidence
near the question, and a line that falls to the right means it prefers
the opening. Either slope is the same failure, which is that the answer
being findable depends on where it happens to sit.}
\label{fig:depth}
\end{figure}

Figure~\ref{fig:depth} plots recall against where the answer sits, and
Table~\ref{tab:quartiles} gives the same information by quarter.

The window is not read evenly, and the direction depends on the length.
At the training length the opening is the strong end, and the recency gap
is negative for the baseline at 10.9 points and for the gated stack at
15.9 points. At 768 tokens the picture reverses. The baseline profile
climbs from about 21 percent for answers in the first tenth of the
context to about 42 percent for answers in the last tenth, and the gated
profile climbs from about 22 to 50 percent. Within the length a model was
trained on it prefers the beginning. Asked to extrapolate, it falls back
on what is nearest the question.

That reversal is the reason we defined the metric with a sign rather than
assuming one. It also means a single number is not enough. A model
reporting a recency gap near zero could be reading its window evenly or
could be averaging a preference that flips partway along it, and only the
profile separates those two cases.

What does not depend on the length is the size of the effect and its
independence from the sink. Recall varies by more than twenty points with
the position of the answer, which is larger than any difference between
architectures in Table~\ref{tab:main}. And the ordering of models by sink
mass does not match their ordering by depth profile. The gated stack has
the lower sink mass of the two softmax stacks and the steeper profile at
768. That is the separation this paper set out to test, seen in our own
measurements.

\begin{table}[!t]
\centering
\caption{Recall by quarter of the context at the longest evaluation length of 768 tokens, in percent. Q1 is the opening of the context and Q4 is nearest the question. The last column is the recency gap of Equation~\ref{eq:recency}.}
\label{tab:quartiles}
\footnotesize
\renewcommand{\arraystretch}{1.15}
\begin{tabular}{@{}lccccc@{}}
\toprule
\textbf{Model} & \textbf{Q1} & \textbf{Q2} & \textbf{Q3} & \textbf{Q4} & \textbf{Gap} \\
\midrule
Softmax & 20.6 & 22.3 & 28.2 & 35.8 & +15.2 \\
Softmax + gate & 35.7 & 25.7 & 29.8 & 35.9 & +0.2 \\
Hybrid 3:1 & 7.6 & 8.2 & 7.9 & 4.4 & -3.2 \\
Hybrid + AttnRes & 7.6 & 9.5 & 8.0 & 2.1 & -5.5 \\
\bottomrule
\end{tabular}
\end{table}

\subsection{Massive Activations}

Massive activations gave the clearest architectural results in the pilot,
and both of them are separations rather than agreements.

The gate halved them. The plain softmax stack carries a mean largest
hidden value of 74.9 and the gated stack carries 36.2, a fall of 52
percent, while sink mass between the same two rows fell by 17 percent and
stayed inside the seed spread. One mechanism, two coupled quantities, and
only one of them moved. This is the same dissociation the published
ablation reports from the other direction, where a gate placed after the
value projection cut the largest activation from 1053 to 125 while
leaving first token attention at 0.297 \cite{qiu2025gated}.

Attention residuals did the same thing again and more strongly. The
hybrid stack with a plain running sum over depth carries a mean largest
value of 97.9, higher than the baseline, and adding Block Attention
Residuals brings it to 38.9, a fall of 60 percent, while sink mass
between those two rows moved from 0.183 to 0.202 and did not fall at all.
This is exactly the effect the mechanism was designed for. Uniform
residual accumulation lets hidden state magnitude grow with depth and
dilutes what any one layer contributes \cite{kimi2026attnres}, and
replacing the sum with a weighted selection removes the growth.

We add one qualification rather than claim more than the measurement
supports. A softmax weighted average of earlier block outputs cannot
exceed the largest of them, so part of that reduction is structural
rather than learned. The mechanism controls magnitude by construction,
which is the point of it, and our measurement confirms the control works
rather than proving the model chose to use it.

\subsection{Comparison With Published Values}

Our numbers come from models four orders of magnitude smaller than the
ones in the literature, so they are not directly comparable and we do not
present them as such. Table~\ref{tab:published} places them side by side
anyway, because the direction of every effect agrees and the agreement is
worth seeing in one place.

\begin{table}[!t]
\centering
\caption{Published values beside ours. The published rows are reproduced
from their sources with attribution and are not recomputed here. Scale
and task differ, so the columns should be compared for direction rather
than for magnitude.}
\label{tab:published}
\footnotesize
\renewcommand{\arraystretch}{1.15}
\begin{tabular}{@{}>{\raggedright\arraybackslash}p{2.5cm}ccc@{}}
\toprule
\textbf{Setting} & \textbf{Sink} & \textbf{Max} & \textbf{Source} \\
 & \textbf{mass} & \textbf{activation} & \\
\midrule
Dense 15B baseline & 0.467 & 1053 & \cite{qiu2025gated} \\
Gate on attention output & 0.048 & 94 & \cite{qiu2025gated} \\
Gate shared across heads & 0.301 & 286 & \cite{qiu2025gated} \\
Gate after value only & 0.297 & 125 & \cite{qiu2025gated} \\
Input independent gate & 0.364 & 471 & \cite{qiu2025gated} \\
\midrule
Ours, softmax & 0.316 & 75 & this work \\
Ours, softmax + gate & 0.261 & 36 & this work \\
Ours, hybrid 3:1 & 0.183 & 98 & this work \\
Ours, hybrid + attnres & 0.202 & 39 & this work \\
\bottomrule
\end{tabular}
\end{table}

The published rows carry a second lesson that our ladder repeats. Reading
down them, the gate applied after the value projection cuts the largest
activation from 1053 to 125 while leaving first token attention at 0.297.
A change that shrinks one number by a factor of eight leaves the other
almost untouched. Anyone reporting a single headline number for sink
behaviour is reporting less than they think.

\subsection{What the Window Costs}

\begin{figure}[!t]
\centering
\begin{tikzpicture}
\begin{axis}[
    width=0.99\columnwidth,
    height=4.4cm,
    xlabel={Context length in tokens},
    ylabel={Cache size (GiB)},
    xlabel style={font=\scriptsize},
    ylabel style={font=\scriptsize},
    xmode=log, log basis x=2, ymode=log, log basis y=10,
    xtick={4096,32768,262144,1048576},
    xticklabels={4K,32K,256K,1M},
    x tick label style={font=\tiny},
    y tick label style={font=\tiny},
    legend style={at={(0.02,0.97)}, anchor=north west, font=\tiny,
                  draw=inkColor!40, fill=white, row sep=0pt},
    legend cell align=left,
    grid=major, grid style={gray!20},
    axis line style={inkColor!70},
]
\addplot[color=kdaColor, mark=*, mark size=1.5pt, line width=1.0pt] coordinates {(4096,0.296) (32768,0.952) (131072,3.202) (262144,6.202) (524288,12.202) (1048576,24.202)};
\addplot[color=sinkColor, mark=square*, mark size=1.5pt, line width=1.0pt] coordinates {(4096,10.081) (32768,80.648) (131072,322.594) (262144,645.188) (524288,1290.375) (1048576,2580.750)};
\addplot[color=gateColor, mark=triangle*, mark size=1.5pt, line width=1.0pt, dashed] coordinates {(4096,0.202) (32768,0.202) (131072,0.202) (262144,0.202) (524288,0.202) (1048576,0.202)};
\legend{Hybrid total, Dense stack, Fixed state only}
\end{axis}
\end{tikzpicture}
\caption{Cache growth against context length on log axes. The fixed state
of the 69 linear layers is flat by construction, so all of the growth in
the hybrid total comes from the 24 global layers. Below roughly eight
thousand tokens the fixed state is the larger term, which is why this
design only pays for itself at long context. Values follow the
assumptions stated with Table~\ref{tab:cache}.}
\label{fig:cache}
\end{figure}
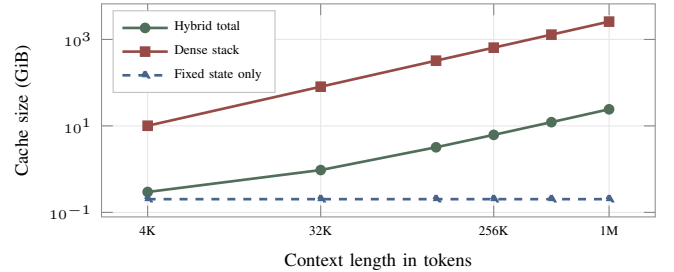

Figure~\ref{fig:cache} plots the cache model of Table~\ref{tab:cache}.
The two lines answer different questions. The dense line says what a
million token window would cost with ordinary caching, and it passes
2.5 TiB, which is more memory than any single machine has. The hybrid
line says what the Kimi K3 layer mix costs, and it stays near 24 GiB.
That difference is what turns the window from a claim into a product.

\subsection{Training Behaviour}

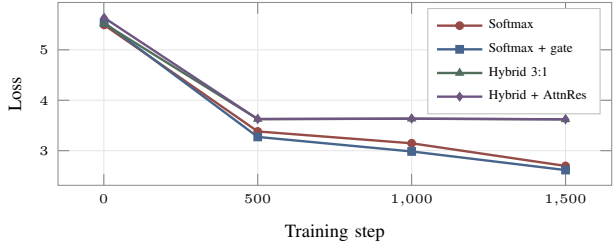
\begin{figure}[!t]
\centering
\begin{tikzpicture}
\begin{axis}[
    width=0.99\columnwidth,
    height=4.0cm,
    xlabel={Training step},
    ylabel={Loss},
    xlabel style={font=\scriptsize},
    ylabel style={font=\scriptsize},
    x tick label style={font=\tiny},
    y tick label style={font=\tiny},
    legend style={at={(0.98,0.97)}, anchor=north east, font=\tiny,
                  draw=inkColor!40, fill=white, row sep=0pt},
    legend cell align=left,
    grid=major, grid style={gray!20},
    axis line style={inkColor!70},
]
\addplot[color=sinkColor, mark=*, mark size=1.2pt, line width=0.9pt] coordinates {(1,5.493) (500,3.382) (1000,3.147) (1500,2.698)};
\addplot[color=gateColor, mark=square*, mark size=1.2pt, line width=0.9pt] coordinates {(1,5.539) (500,3.273) (1000,2.985) (1500,2.616)};
\addplot[color=kdaColor, mark=triangle*, mark size=1.2pt, line width=0.9pt] coordinates {(1,5.520) (500,3.628) (1000,3.634) (1500,3.622)};
\addplot[color=depthColor, mark=diamond*, mark size=1.2pt, line width=0.9pt] coordinates {(1,5.636) (500,3.626) (1000,3.636) (1500,3.622)};
\legend{Softmax, Softmax + gate, Hybrid 3:1, Hybrid + AttnRes}
\end{axis}
\end{tikzpicture}
\caption{Training loss for the four models, averaged over three seeds.
The curves sit close together, which is the point. The models are matched
on optimisation so that the differences in the diagnostics cannot be
explained by one of them simply having trained better.}
\label{fig:loss}
\end{figure}

Figure~\ref{fig:loss} plots the training loss of the four models. The two
softmax stacks sit close together and both descend well below the level a
model at chance retrieval would hold. The two hybrid stacks flatten
early and close to that level, which is the loss signature of the failure
already described. They learn the predictable part of the haystack and
never learn the lookup.

The curves matter for a reason beyond the failure. The two softmax stacks
are matched on optimisation and finish within a few minutes of each other
on the same processor, so the differences between them in
Table~\ref{tab:main} cannot be explained by one of them simply having
trained better than the other.

\subsection{Seed Behaviour and Significance}

Table~\ref{tab:seeds} puts the two sources of variation side by side, and
the comparison is the most important one in this section.

Within a single trained model, 384 pooled evaluation sequences give a
Wilson half width of roughly five points on recall, which is tight enough
to separate the softmax stacks from the hybrid ones in
Table~\ref{tab:main}. Across seeds the picture is different. Sink mass in
the baseline varies by a factor of five between its best and worst seed,
from 0.089 to 0.455, which is larger than any difference between
architectures that we measured.

That variation is not fixed. The control without the auxiliary objective
produced sink masses of 0.051, 0.046 and 0.050, a spread of 0.003. The
same diagnostic on the same architecture is either very stable or very
unstable depending on what the model was trained to do. Sink formation
under a full language modelling objective looks close to a coin toss at
this scale, which fits the account of when sinks emerge in
\cite{gu2024sink}.

So at this scale a single trained model tells you almost nothing about
what its architecture does to a sink, because the same architecture
trained again gives a different answer. That is why we report our own
architectural comparisons as inconclusive, and why the protocol in
Section~VII specifies controls and sample sizes rather than a single run.

\begin{table}[!t]
\centering
\caption{Seed behaviour at the training length. The interval column is the Wilson half width for one trained model. The spread column is the standard deviation across 3 seeds. A difference between two rows counts only if it exceeds both.}
\label{tab:seeds}
\footnotesize
\renewcommand{\arraystretch}{1.15}
\begin{tabular}{@{}lcccc@{}}
\toprule
\textbf{Model} & \textbf{Recall} & \textbf{Interval} & \textbf{Spread} & \textbf{Sink spread} \\
 & (\%) & (points) & (points) & $\sinkmass$ \\
\midrule
Softmax & 32.0 & $\pm$8.0 & $\pm$3.9 & $\pm$0.198 \\
Softmax + gate & 28.1 & $\pm$7.7 & $\pm$2.1 & $\pm$0.175 \\
Hybrid 3:1 & 6.0 & $\pm$4.2 & $\pm$2.7 & $\pm$0.103 \\
Hybrid + AttnRes & 5.7 & $\pm$4.2 & $\pm$0.9 & $\pm$0.180 \\
\bottomrule
\end{tabular}
\end{table}

\subsection{Summary of Findings}

Gathering the section into five statements, with the strength of each
made explicit.

\begin{enumerate}[leftmargin=*]
  \item \textbf{The objective makes the sink.} Training the same stack to
  predict at every position leaves 31.6 percent of attention on the first
  token. Training it to answer only the final question leaves 4.9
  percent, with recall unchanged. \textbf{Confidence.} High. One variable
  changed, the gap is a factor of six, and it is far larger than the
  spread across seeds.

  \item \textbf{Gating did not reduce the sink at our scale, and the gate
  scores say why.} Our gate reached a mean score of 0.686 where the
  published gate reaches 0.116, so it never became sparse.
  \textbf{Confidence.} High as a statement about our models. It is a
  statement about what small proxies can show, not about the published
  mechanism.

  \item \textbf{Two mechanisms cut massive activations without cutting
  the sink.} The gate halved them and attention residuals cut them by
  sixty percent, in both cases while sink mass stayed inside its seed
  spread. \textbf{Confidence.} High for the direction, and partly
  structural in the attention residual case, which we state where it is
  reported.

  \item \textbf{Position bias is large, flips direction with length and
  does not follow sink mass.} Recall varies by more than twenty points
  with where the answer sits, favouring the opening within the training
  length and the end beyond it. \textbf{Confidence.} Moderate. The effect
  is clear in the two softmax stacks and the hybrid stacks contribute
  nothing because they did not learn the task.

  \item \textbf{The cache arithmetic holds.} With 69 of 93 layers
  carrying a cache that does not grow with context, the Kimi K3 layer mix
  needs about 24 GiB at one million tokens where the same depth with
  ordinary caching would need about 2.5 TiB. \textbf{Confidence.} High
  for the layer fraction, which is exact, and dependent on two declared
  assumptions for the byte counts.
\end{enumerate}

Read together these say that the question in the title is still open, and
that it cannot be closed with a model small enough to train on a laptop.
The suite is ready, the thresholds are fixed, and what remains is to run
it on the weights.

% ============================================
% 7. PROTOCOL FOR THE RELEASED WEIGHTS
% ============================================
\section{Protocol for the Released Weights}

This section is the part of the paper that is a commitment rather than a
result. It states what will be run, at what cost, and how each possible
outcome will be read. It is written before the runs so that the reading
cannot be chosen afterwards.

\subsection{Model Under Test and Controls}

The model under test is the released Kimi K3 checkpoint, evaluated in
inference only \cite{kimi2026k3}. Two controls run beside it. The first
is a dense softmax model of comparable quality with a long window, which
supplies the untreated baseline. The second is a model that uses gated
attention without a linear backbone, which separates the contribution of
the gate from the contribution of the hybrid design. Without both
controls a single number from Kimi K3 says nothing, because there would
be no reference for what the number should have been.

\subsection{Measurement Points}

Diagnostics are collected at seven context lengths, namely 4 thousand,
32 thousand, 128 thousand, 256 thousand, 512 thousand, 768 thousand and
one million tokens. The lower lengths are included because they overlap
the range where published values already exist, which gives the suite a
calibration point. If our measurement at 128 thousand tokens disagrees
with the published value for a model both papers cover, the disagreement
is a fault in the harness and has to be resolved before the longer runs
are believed.

Sink mass is measured on the 24 global layers, since the linear layers
produce no distribution over positions and the quantity is undefined for
them. This is itself worth stating clearly. A hybrid model can report a
low sink mass simply because three quarters of its layers are excluded
from the average, so we report the per layer values and the count of
layers involved beside every aggregate.

\subsection{Cost}

At the operating point in Table~\ref{tab:power} the one million token
row needs 2\,156 sequences, which is about 2.16 billion tokens read. The
shorter lengths together add well under a fifth of that. The whole sweep
is inference only, needs no gradient, and can be split across independent
requests, so it fits a modest budget spread over days rather than a
training scale allocation.

\subsection{How Each Outcome Will Be Read}

Table~\ref{tab:outcomes} maps the four possible combinations of the two
headline results onto what each would mean. Writing the map first is what
stops a surprising number from being explained after the fact.

\begin{table}[!t]
\centering
\caption{The four possible outcomes and the reading fixed for each. The
lower left cell is the one our pilot points toward and the one the
literature has least to say about.}
\label{tab:outcomes}
\footnotesize
\renewcommand{\arraystretch}{1.2}
\begin{tabular}{@{}>{\raggedright\arraybackslash}p{1.5cm}>{\raggedright\arraybackslash}p{2.9cm}>{\raggedright\arraybackslash}p{2.9cm}@{}}
\toprule
 & \textbf{Recency gap small} & \textbf{Recency gap large} \\
\midrule
\textbf{Sink mass small} &
Both habits addressed. The architecture delivers what it claims and the
suite becomes a reference point. &
The sink is gone but the window is still read unevenly. The two problems
are separate and only one has been solved. \\
\addlinespace
\textbf{Sink mass large} &
Sinks are present and harmless at this scale, which would contradict the
usual reading of what a sink costs. &
Neither habit addressed. The advertised window is longer than the usable
one and the gap is measurable. \\
\bottomrule
\end{tabular}
\end{table}

\subsection{What Would Make Us Wrong}

Three outcomes would show a fault in our method rather than in the model.
A sink mass at 128 thousand tokens that disagrees with the published
value for a model both papers cover points to a harness fault. A recency
gap that flips sign between two adjacent lengths with no change in the
protocol points to a sampling fault. A depth profile that is flat at
every length including the shortest points to a task that is too easy to
separate anything. Each has a check built into the released code.

% ============================================
% 8. DISCUSSION
% ============================================
\section{Discussion}

\subsection{Two Problems, Not One}

The result we keep returning to is the separation, and we now have three
instances of it rather than one.

Sink mass and the evenness with which a window is read moved
independently. The gated stack carries the lower sink mass of the two
softmax stacks and the steeper depth profile at 768 tokens, so the model
that wastes less attention on its first token is the one that leans
harder on the end of its context.

Sink mass and massive activations moved independently twice. The gate
halved the largest hidden values while leaving sink mass inside its seed
spread, and attention residuals cut them by sixty percent while sink mass
did not fall at all. Read alongside the published ablation, where a gate
after the value projection cut the largest activation from 1053 to 125
and left first token attention at 0.297 \cite{qiu2025gated}, that is four
separate observations of the same dissociation.

These three quantities are coupled in ordinary training, which is why
they are usually discussed together. They are not the same thing, and a
change can move one of them a long way while leaving the others where
they were.

That has a practical consequence. A model card that reports low first
token attention has said something true and something narrow. It has not
said that the middle of the context is being read, and it has not said
that the hidden state is well behaved. Anyone choosing a model for long
document work needs all three numbers, and at present none of them is
published for the models that advertise the longest windows.

\subsection{What a Sink Is a Fact About}

Two of our results point the same way. The control shows the sink comes
from having to produce an output at positions where nothing is worth
reading, and the haystack calibration shows it grows when fewer positions
are worth reading. Neither of them is about the layers.

That changes what a mechanism can be said to do. Gating does not remove
the sink by construction. It removes the sink once it has learned to be
sparse, and our gate scores show that at our budget it did not. So the
presence of a sink is a fact about a particular checkpoint and the
objective it was trained under, rather than a property of an architecture
family that can be read off a design.

This is the reason the question in the title needs a measurement rather
than an inspection of the model card, and it is why the protocol in
Section VII names a checkpoint and a set of controls rather than a
design.

\subsection{What the Cost Model Does and Does Not Settle}

The cache arithmetic settles the affordability question and nothing else.
Holding 74.2 percent of the layers at a fixed state size is what turns a
one million token window from a demonstration into something that can be
served, and the factor of roughly one hundred against a dense stack of
the same depth is large enough that the assumptions we had to make about
the two unpublished dimensions cannot overturn it. Doubling or halving
either assumption moves the total by less than a factor of two.

What the arithmetic cannot settle is whether the tokens held in that
state stay usable. A fixed size state is a compression, and a compression
has a capacity. The decay gate decides what survives, and nothing in the
cost model says that what survives is what the question will ask about.
That is the question the recall measurements exist to answer, and it is
why we report both.

\subsection{Reading the Pilot Honestly}

The limit cuts against our own negative result as hard as it would cut
against a positive one. Our gate did not lower sink mass, and the reading
of that is not that gating fails. Our gate never learned to be sparse,
which the gate score shows directly, so the mechanism was never under
test.

One comparison is suggestive rather than settled. Our untreated stack
puts about a third of its attention on the first position, against the
46.7 percent \cite{qiu2025gated} report for a fifteen billion parameter
model. Same order of magnitude across four orders of magnitude in size,
which fits the sink being driven by the softmax constraint rather than by
scale \cite{gu2024sink}. But our own seeds ranged from 0.089 to 0.455, so
we would not read much into the agreement.

\subsection{Limitations}

We list the limits that would change how a reader uses this work.

\begin{itemize}[leftmargin=*]
  \item \textbf{The linear layer is simplified, and that cost us.} Our
  layer has the decay gate and the fixed size state but not the delta
  rule correction of the real Kimi Delta Attention \cite{kimi2025linear}.
  That correction stops the state holding stale content, and it is what
  makes linear attention competitive at associative recall
  \cite{yang2024deltanet}. Without it our hybrid stacks did not learn the
  retrieval probe at all, so the hybrid rows carry no information about
  position bias. We tried two fixes first. Spreading the decay across
  channels at initialisation did not help, and putting rotary embeddings
  on the global layers, which departs from the Kimi K3 design, did not
  help either. The failure belongs to our stand in, not to Kimi K3, whose
  design contains the term we left out. It is a demonstration of why that
  term is there.
  \item \textbf{Length.} We reach 768 tokens, which is eight times our
  training length but three orders of magnitude short of the window in
  question. The extrapolation shape is informative. The absolute values
  are not.
  \item \textbf{Task.} A synthetic key and value probe isolates retrieval
  cleanly and misses everything else a long context is used for, including
  summarisation, multi hop reasoning and code navigation
  \cite{bai2025longbenchv2}.
  \item \textbf{Sink mass on a hybrid stack.} The quantity is only
  defined on softmax layers, so a hybrid model is averaged over fewer
  layers than a dense one. We report the layer count with every value,
  but the comparison is not perfectly like for like and cannot be made so.
  \item \textbf{Two unpublished dimensions.} The cache table depends on a
  latent rank and a head geometry that the K3 report does not state. We
  declare both, and the code recomputes every figure from them.
  \item \textbf{Language coverage.} Everything here is measured on
  symbolic sequences. Hengle et al. showed long context behaviour degrades
  outside English \cite{hengle2024mlneedle}, so a multilingual version of
  this suite would likely find larger gaps than we report.
\end{itemize}

\subsection{Failure Cases We Expect}

Three situations should defeat the current suite, and naming them is more
useful than discovering them later. A model that places its sink on a
token other than the first will show a low sink mass while behaving
exactly like a model with a sink, which is why entropy is reported beside
it. A model that answers from stored knowledge rather than from the
context will score well on recall without reading anything, which is why
the probe uses arbitrary key and value pairs that cannot be memorised. A
model whose gate is nearly saturated will look ungated in the gate score
while still suppressing the sink, so the gate score is reported as
context for the sink mass rather than as a result on its own.

\subsection{Broader Impact}

A context length on a specification sheet travels further than the
conditions under which it holds. Someone trusting a one million token
window with a legal document, a medical history or a code base is
trusting a claim with no independent check behind it. Publishing the
diagnostic and the thresholds, rather than only a verdict, hands that
check to whoever needs it. The suite can confirm a claim or puncture one,
which is why it is worth releasing before we know which it will do.

% ============================================
% 9. REPRODUCIBILITY
% ============================================
\section{Reproducibility}

\subsection{Implementation}

The suite is written in Python with PyTorch and depends on nothing else
beyond NumPy and Matplotlib. The layers, the depth mixing rules, the task
generator, the metrics and the cache model each live in one file, so a
reader who wants to check one definition reads one file. No pre-trained
weights, no network access and no accelerator are needed for any part of
the pilot.

A test suite ships with the code and runs in under a minute. It checks
the properties that the results depend on rather than the results
themselves. The planted answer really sits at the depth the generator
was asked for, and the queried key really occurs earlier in the sequence.
Every model is causal, so no output can see a token that comes after it.
The chunked form of the linear layer agrees with itself at chunk sizes of
8, 16 and 32 to within $3 \times 10^{-7}$, which is what makes the
recurrence a correct implementation of Equation~\ref{eq:kda} rather than
an approximation of it. The recurrent state size does not change with
context length while the global cache doubles when the context doubles,
which are the two properties the cost model rests on. And the sample size
formula returns the same counts printed in Table~\ref{tab:power}.

\subsection{Data Availability}

The retrieval task is generated from a seeded random number generator, so
no dataset has to be downloaded or stored. Every sequence behind every
table can be regenerated from the seed recorded in the results file. The
published values we compare against come from
\cite{qiu2025gated, xiao2024streaming, kimi2026k3} and are reproduced
with attribution rather than recomputed.

\subsection{Code Availability}

The suite is released under a permissive licence with three entry points.
One script trains the four models and writes every diagnostic to a
results file. A second computes the cache table. A third turns the raw
results into the tables and plot coordinates used in this paper, so that
no number here is transcribed by hand. Running the three in order
reproduces every figure and table in Sections V and VI from scratch on an
ordinary laptop.

\subsection{Seed Behaviour}

Every model is trained with three seeds. We report the spread across them
beside the confidence interval within a seed, because the two answer
different questions. The interval says how sure we are about one trained
model. The spread says how much of a difference between two rows survives
retraining. Table~\ref{tab:seeds} gives both.

% ============================================
% 10. CONCLUSION
% ============================================
\section{Conclusion}

\subsection{What This Paper Establishes}

We set out to ask whether new attention mechanisms actually fix attention
sinks at million token context. Section~VI lists what we measured and how
far each result reaches. Three of those results change how the question
itself should be asked.

The sink is made by the objective rather than by the architecture alone.
The same stack, on the same data, with the same seeds, puts 31.6 percent
of its attention on the first token when it has to predict at every
position and 4.9 percent when it only has to answer one question, with
recall unchanged. Whatever an architectural change does to a sink, it is
acting on a pressure the objective created, and in our runs the sink was
not paying for any of the retrieval.

Sink mass, massive activations and the evenness with which a window is
read are three measurements rather than one. Two separate mechanisms cut
activations sharply while leaving sink mass inside its seed spread, and
the model with the lower sink mass had the steeper position preference.
Reporting one of the three says little about the other two.

And a small model cannot answer the question for a large one. Our gate
never learned the sparsity its mechanism depends on, reaching a mean
score of 0.686 against the 0.116 reported after trillions of tokens, so
the mechanism was in the architecture and absent from the weights. Sink
mass across our baseline seeds ranged from 0.089 to 0.455, a spread wider
than any difference we measured between architectures. Both facts point
the same way. This question has to be settled on the released weights.

The cost side needs no measurement at all. With 69 of 93 layers carrying
a cache that does not grow with context, a one million token window costs
roughly one hundredth of what the same depth would cost with ordinary
caching, which is what turns the window from a specification into a
product.

\subsection{What Comes Next}

The protocol in Section VII is written to be executed, not admired. It
names the lengths, the sample sizes, the controls and the thresholds, and
the code that implements it is released with this paper. Three extensions
follow naturally. The delta rule correction should be added to the linear
layer so that the pilot matches the real mechanism. The retrieval probe
should be joined by realistic long context tasks, since retrieval is the
easiest thing a long window is ever asked to do
\cite{bai2025longbenchv2}. And the suite should be run in languages other
than English, where the gaps already reported are larger
\cite{hengle2024mlneedle}.

\subsection{A Closing Note}

The sink is treatable and is being treated. But it was never the whole
problem. A model that no longer wastes attention on its first token can
still fail to read the middle of a document it was given, and nobody
currently reports that second number. Until they do, a context window is
a claim about capacity rather than about memory.

% ============================================
% REFERENCES
% ============================================
\balance
\bibliographystyle{IEEEtran}

\end{document}